\documentclass[11pt, logo, onecolumn,letterpaper]{googledeepmind}

\usepackage[authoryear, round]{natbib}
\usepackage{hyperref}

\hypersetup{
    citecolor = {cyan},
    linkcolor = {magenta}
}

\usepackage{microtype}
\usepackage{graphicx}
\usepackage{booktabs} %
\usepackage{float}

\usepackage{amsmath}
\usepackage{amssymb}
\usepackage{mathtools}
\usepackage{amsthm}
\usepackage{mathrsfs}
\usepackage{empheq}
\usepackage{nicefrac}
\usepackage{dsfont}
\usepackage{enumitem}
\usepackage[frozencache=true,cachedir=minted-cache]{minted}
\usepackage{float}
\usepackage{subcaption}
\usepackage{adjustbox}
\usepackage{threeparttable}
\usepackage{booktabs}
\usepackage{wrapfig}

\setminted[python]{frame=lines, breaklines, framesep=2mm, fontsize=\scriptsize}

\definecolor{blanchedalmond}{rgb}{1.0, 0.92, 0.8}
\definecolor{carmine}{rgb}{0.59, 0.0, 0.09}
\definecolor{lightblue}{rgb}{0.22,0.45,0.70}%

\renewcommand{\mathbf}{\boldsymbol}

\makeatletter
\def\Ddots{\mathinner{\mkern1mu\raise\p@
\vbox{\kern7\p@\hbox{.}}\mkern2mu
\raise4\p@\hbox{.}\mkern2mu\raise7\p@\hbox{.}\mkern1mu}}
\makeatother

\numberwithin{equation}{section}

\definecolor{amaranth}{rgb}{0.9, 0.17, 0.31}
\definecolor{antiquebrass}{rgb}{0.8, 0.58, 0.46}
\definecolor{antiquefuchsia}{rgb}{0.57, 0.36, 0.51}
\definecolor{chromeyellow}{rgb}{0.31, 0.47, 0.26}

\newcommand{\hlgauss}{HL-Gauss}
\newcommand{\twohot}{Two-Hot}

\DeclareMathOperator*{\argmax}{arg\,max}
\DeclareMathOperator*{\argmin}{arg\,min}

\newcommand{\qapprox}{Q}
\newcommand{\zapprox}{Z}
\newcommand{\ind}{\mathds{1}}
\newcommand{\tdtarget}{(\widehat{\mathcal{T}} \qapprox)}

\newcommand{\cdrl}{CDRL}
\DeclarePairedDelimiter\ceil{\lceil}{\rceil}
\DeclarePairedDelimiter\floor{\lfloor}{\rfloor}

\makeatletter
\renewcommand*{\appendixautorefname}{\S\@gobble}
\renewcommand*{\sectionautorefname}{\S\@gobble}
\renewcommand*{\subsectionautorefname}{\S\@gobble}
\makeatother

\newcommand{\sref}[1]{{\color{magenta}\S}\ref{#1}}

\title{{Stop Regressing}: Training Value Functions via Classification for Scalable Deep RL}

\correspondingauthor{jfarebro@cs.mcgill.ca, aviralkumar@google.com, rishabhagarwal@google.com}

\renewcommand{\today}{}

\author[1,2,*]{Jesse Farebrother}
\author[1,$\dagger$]{Jordi Orbay}
\author[1,$\dagger$]{Quan Vuong}
\author[1,$\dagger$]{Adrien Ali Ta\"{i}ga}
\author[1]{Yevgen Chebotar}
\author[1]{Ted Xiao}
\author[1]{Alex Irpan}
\author[1]{Sergey Levine}
\author[1,3,$\dagger$]{Pablo Samuel Castro}
\author[1]{Aleksandra Faust}
\author[1,$\dagger$]{Aviral Kumar}
\author[1,3,*]{Rishabh Agarwal}

\affil[*]{Equal Contribution}
\affil[$\dagger$]{Core Contribution}
\affil[1]{Google DeepMind}
\affil[2]{Mila, McGill University}
\affil[3]{Mila, Universit\'e de Montr\'eal}

\begin{abstract}
Value functions are a central component of deep reinforcement learning (RL). These functions, parameterized by neural networks, are trained using a mean squared error regression objective to match bootstrapped target values.
However, scaling value-based RL methods that use regression to large networks, such as high-capacity Transformers, has proven challenging. %
This difficulty is in stark contrast to supervised learning: by leveraging a cross-entropy classification loss, supervised methods have scaled reliably to massive networks. Observing this discrepancy, in this paper, we investigate whether the scalability of deep RL can also be improved simply by using classification in place of regression for training value functions. We demonstrate that value functions trained with categorical cross-entropy significantly improves performance and scalability in a variety of domains. These include: single-task RL on Atari 2600 games with SoftMoEs, multi-task RL on Atari with large-scale ResNets, robotic manipulation with Q-transformers, playing Chess without search, and a language-agent Wordle task with high-capacity Transformers, achieving \emph{state-of-the-art results} on these domains.
Through careful analysis, we show that the benefits of categorical cross-entropy primarily stem from its ability to mitigate issues inherent to value-based RL, such as noisy targets and non-stationarity. 
Overall, we argue that a simple shift to training value functions with categorical cross-entropy can yield substantial improvements in the scalability of deep RL at little-to-no cost.
\end{abstract}

\begin{document}

\maketitle

\section{Introduction}\label{introduction}
\vspace{-0.2cm}

A clear pattern emerges in deep learning breakthroughs -- from AlexNet~\citep{krizhevsky2012imagenet} to Transformers~\citep{vaswani2017attention} -- classification problems seem to be particularly amenable to effective training with large neural networks.
Even in scenarios where a regression approach appears natural, framing the problem instead as a classification problem often improves performance~\citep{torgo1996regression, rothe2018deep, rogez2019lcr}.
This involves converting real-valued targets into categorical labels and minimizing categorical cross-entropy rather than the mean-squared error.
Several hypotheses have been put forward to explain the superiority of this approach, including 
stable gradients~\citep{imani2018improving,imani2024investigating},
better representations~\citep{zhang2023improving}, implicit bias~\citep{stewart2023regression}, and dealing with imbalanced data~\citep{pintea2023step} -- suggesting their potential utility beyond supervised regression.

Unlike trends in supervised learning, value-based reinforcement learning~(RL) methods primarily rely on regression. For example, deep RL methods such as deep Q-learning~\citep{mnih15dqn} and actor-critic~\citep{mnih2016asynchronous} use a regression loss, such as mean-squared error, to train a value function from continuous scalar targets. While these value-based deep RL methods, powered by regression losses, have led to high-profile results~\citep{silver2017mastering}, it has been challenging to scale them up to large networks, such as high-capacity transformers. This lack of scalability has been attributed to several issues~\citep{kumar2021implicit, kumar2021dr3, agarwal2021precipice, lyle2022understanding, lelan23bootstrap, obando2024mixtures}, but
\emph{\textbf{what if simply reframing the regression problem as classification can enable the same level of scalability achieved in supervised learning?}}

\begin{figure*}[t]
    \centering
    \includegraphics[width=0.95\linewidth]{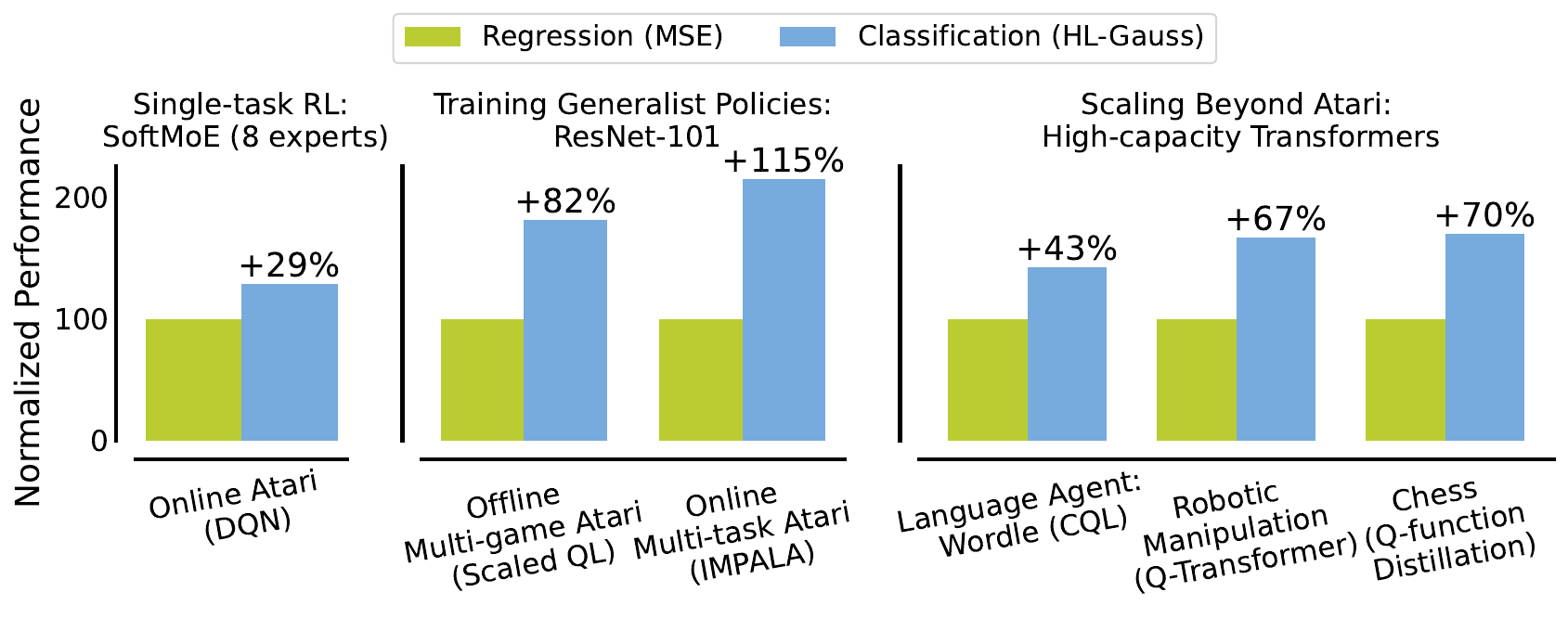}
    \vspace{-0.1cm}
    \caption{\footnotesize{\textbf{Performance gains from \hlgauss{} cross-entropy loss~(\sref{subsec:transform_q}) over MSE} regression loss for training value-networks with modern architectures, including MoEs~(\sref{moe}), ResNets~(\sref{sec:generalist}), and Transformers~(\sref{sec:beyond_atari}). The x-axis labels correspond to domain name, with training method in brackets. For multi-task RL results, we report gains with ResNet-101 backbone, the largest network in our experiments. For Chess, we report improvement in performance gap relative to the teacher Stockfish engine, for the 270M transformer. For Wordle, we report results with behavior regularization of 0.1.}
    \label{fig:topline_figure}}
\end{figure*}

In this paper, we perform an extensive study to answer this question by assessing the efficacy of various methods for deriving classification labels for training a value-function with a categorical cross-entropy loss. Our findings reveal that training value-functions with cross-entropy substantially improves the performance, robustness, and scalability of deep RL methods (\autoref{fig:topline_figure}) compared to traditional regression-based approaches. The most notable method \citep[\hlgauss{};][]{imani2018improving} leads to consistently 30\% better performance when scaling parameters with Mixture-of-Experts in single-task RL on Atari~\citep{obando2024mixtures}; $\mathbf{1.8-2.1\times}$ performance in multi-task setups on Atari~\citep{kumar2022offline, taiga2023multitask}; $\mathbf{40}$\% better performance in the language-agent task of Wordle~\citep{snell2022offline}; $\mathbf{70}$\% improvement for playing chess without search~\citep{ruoss2024grandmaster}; and $\mathbf{67}$\% better performance on large-scale robotic manipulation with transformers~\citep{chebotar2023q}. The consistent trend across diverse domains, network architectures, and algorithms highlights the substantial benefits of treating regression as classification in deep RL, underscoring its potential as a pivotal component as we move towards scaling up value-based RL.

With \textbf{strong empirical results to support the use of cross-entropy as a ``drop-in'' replacement for the mean squared error~(MSE) regression loss in deep RL}, we also attempt to understand the source of these empirical gains. Based on careful diagnostic experiments, we show that the categorical cross-entropy loss offers a number of benefits over mean-squared regression.
Our analysis suggests that the categorical cross-entropy loss mitigates several issues inherent to deep RL, including robustness to noisy targets and allowing the network to better use its capacity to fit non-stationary targets.
These findings not only help explain the strong empirical advantages of categorical cross-entropy in deep RL but also provide insight into developing more effective learning algorithms for the field.

\vspace{-0.2cm}
\section{Preliminaries and Background}\label{sec:rl_back}
\vspace{-0.2cm}
\begin{figure*}[t]
\centering
\includegraphics[width=0.975\linewidth]{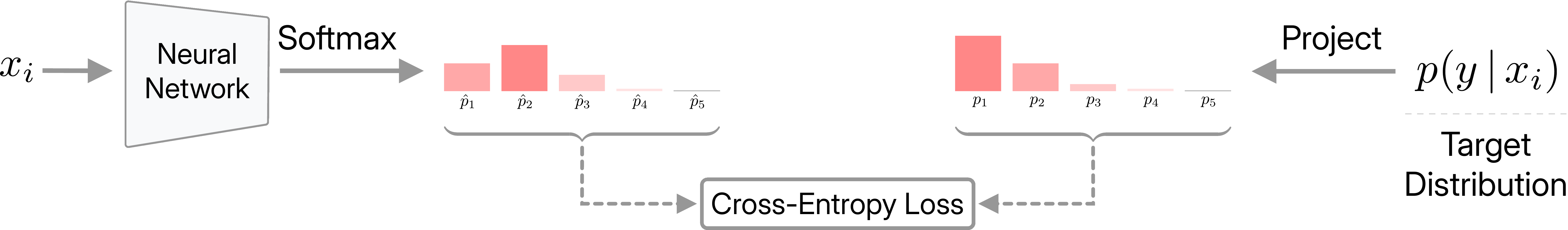}
\caption{\footnotesize\textbf{Regression as Classification.} Data points $\mathbf{x}_i$ are transformed by a neural network to produce a categorical distribution via a softmax. The prediction $\hat{y}$ is taken to be the expectation of this categorical distribution. The logits of the network are reinforced by gradient descent on the cross-entropy loss with respect to a target distribution whose mean is the regression target $y_i$. \autoref{fig:methods} depicts three methods for constructing and projecting the target distribution in RL.}
\vspace{-0.05cm}
\label{fig:reg_as_class}
\end{figure*}

\textbf{Regression as classification.} We take a probabilistic view on regression where given input $x \in \mathbb{R}^d$ we seek to model the target as a conditional distribution $Y \,|\, x \sim \mathcal{N}(\mu = \hat{y}(x; \theta), \sigma^2)$ for some fixed variance $\sigma^2$ and predictor function $\hat{y} : \mathbb{R}^d \times \mathbb{R}^{k} \to \mathbb{R}$ parameterized by the vector $\theta \in \mathbb{R}^k$.
The maximum likelihood estimator for data $\{ x_{i}, y_i \}_{i=1}^N$ is characterized by the \textbf{mean-squared error~(MSE)} objective,
\begin{align*}
\min_{\theta} \, \sum_{i=1}^N \left( \hat{y}(x_i; \theta) - y_i \right)^2 \,,
\end{align*}
with the optimal predictor being $\hat{y}(x; \theta^\ast) = \mathbb{E} \left[ Y \,|\, x \right]$.

Instead of learning the mean of the conditional distribution directly, an alternate approach is to learn a distribution over the target value, and then, recover the prediction $\hat{y}$ as a statistic of the distribution.
To this end, we will construct the target distribution $Y \,|\, x$ with probability density function $p(y \,|\, x)$ such that our scalar target can be recovered as the mean of this distribution $y = \mathbb{E}_p \left[ Y \,|\, x \right]$. We can now frame the regression problem as learning a parameterized distribution $\hat{p}(y\,|\,x; \theta)$ that minimizes the KL divergence to the target $p(y\,|\,x)$,
{
\begin{align}
\min_{\theta} \sum_{i=1}^N  \int_{\mathcal{Y}} p(y\,|\, x_{i}) \log{(\hat{p}(y \,|\, x_i; \theta))} \, dy\label{eq:kl}
\end{align}}%
which is the cross-entropy objective.
Finally, our prediction can be recovered as $\hat{y}(x; \theta) = \mathbb{E}_{\hat{p}} \left[\, Y \,|\, x; \theta \,\right]$.

Given this new problem formulation, in order to transform the distribution learning problem into a tractable loss we restrict $\hat{p}$ to the set of categorical distributions supported on $[v_{\text{min}}, v_{\text{max}}]$ with $m$ evenly spaced locations or ``classes'', $v_{\text{min}} \le z_1 < \cdots < z_m \le v_{\text{max}}$ defined as,
\begin{equation}
\mathcal{Z} = \left\{ \sum_{i=1}^m p_i \,\delta_{z_i} \,:\, p_i \ge 0, \sum_{i=1}^m p_i = 1 \right\} \, ,
\label{eq:cat}
\end{equation}
where $p_i$ is the probability associated with location $z_i$ and $\delta_{z_i}$ is the Dirac delta function at location $z_i$.
The final hurdle is to define a procedure to construct the target distribution $Y \,|\, x$ and its associated projection onto the set of categorical distributions $\mathcal{Z}$.
We defer this discussion to \sref{sec:td_target} where we discuss various methods for performing these steps in the context of RL.

\textbf{Reinforcement Learning~(RL).} We consider the reinforcement learning (RL) problem where an agent interacts with an environment by taking an action $A_t \in \mathcal{A}$ in the current state $S_t \in \mathcal{S}$ and subsequently prescribed a reward $R_{t+1} \in \mathbb{R}$ before transitioning to the next state $S_{t+1} \in \mathcal{S}$ according to the environment transition probabilities.
The return numerically describes the quality of a sequence of actions as the cumulative discounted sum of rewards $G_t = \sum_{k=0}^\infty \gamma^k R_{t+k+1}$ where $\gamma \in [0, 1)$ is the discount factor. The agent's goal is to learn the policy $\pi: \mathcal{S} \to \mathscr{P}(\mathcal{A})$ that maximizes the expected return.
The action-value function allows us to query the expected return from taking action $a$ in state $s$ and following policy $\pi$ thereafter: $q_\pi(s, a) = \mathbb{E}_{\pi} \left[ G_t \,|\, S_t = s,\, A_t = a \right]$.

Deep Q-Networks \citep[DQN;][]{mnih15dqn} proposes to learn the approximately optimal state-action value function $\qapprox(s, a; \theta) \approx q_{\pi^\ast}(s, a)$ with a neural network parameterized by $\theta$. Specifically, DQN minimizes the mean-squared temporal difference (TD) error from transitions $(S_t, A_t, R_{t+1}, S_{t+1})$ sampled from dataset $\mathcal{D}$,
{\begin{empheq}[box=\fbox]{align}\label{eq:tdmse}
\text{TD}_{\text{MSE}}(\theta) = 
\mathbb{E}_{\mathcal{D}} \left[ \left( \tdtarget(S_t, A_t; \theta^{-}) - \,\qapprox(S_t, A_t; \theta) \right)^2 \right]
\end{empheq}
}
where $\theta^{-}$ is a slow moving copy of the parameters $\theta$ that parameterize the ``target network'' and
$$
\tdtarget(s, a; \theta^{-}) = R_{t+1} + \gamma \max_{a'} \qapprox(S_{t+1}, a'; \theta^{-}) \,\,\big|\,\, S_t = s, \, A_t = a \,,
$$
is the sample version of the Bellman optimality operator which defines our scalar regression target.
Most deep RL algorithms that learn value functions use variations of this basic recipe, notably regressing to predictions obtained from a target value network. 

In addition to the standard online RL problem setting, we also explore the offline RL setting where we train agents using a fixed dataset of environment interactions~\citep{agarwal2020optimistic,levine2020offline}. One widely-used offline RL method is CQL~\citep{kumar2020conservative} that jointly optimizes the TD error with a behavior regularization loss with strength $\alpha$, using the following training objective: 
{
\begin{align}
    \min_{\theta}~ \alpha\, \bigg( \mathbb{E}_{\mathcal{D}} \left[\log \big(\sum_{a'} \exp(\qapprox(S_{t+1}, a'; \theta)) \big) \right]\, - \mathbb{E}_{\mathcal{D}}\left[\qapprox(S_t, A_t; \theta)\right]  \bigg) + \text{TD}_{\text{MSE}}(\theta),
\end{align}}%
This work aims to replace the fundamental mean-squared TD-error objective with a classification-style cross-entropy loss for both value-based and actor-critic methods, in both offline and online domains.

\section{Value-Based RL with Classification}\label{sec:td_target}

In this section, we describe our approach to cast the regression problem appearing in TD-learning as a classification problem.
Concretely, instead of minimizing the squared distance between the scalar Q-value and its TD target (\autoref{eq:tdmse}) we will instead minimize the distance between categorical distributions representing these quantities. 
To employ this approach, we will first define the categorical representation for the action-value function $\qapprox(s, a)$.

\noindent\textbf{Categorical Representation.} We choose to represent $\qapprox$ as the expected value of a categorical distribution $\zapprox \in \mathcal{Z}$. This distribution is parameterized by probabilities $\hat{p}_i(s, a; \theta)$ for each location or ``class'' $z_i$ which are derived from the logits $l_i(s, a; \theta)$ through the softmax function:
\begin{align*}
    \qapprox(s, a; \theta) = \mathbb{E}\left[\, \zapprox(s, a; \theta) \,\right], \,\;~~~~~ 
    \zapprox(s, a; \theta) = \sum_{i=1}^m \hat{p}_i(s,a; \theta) \cdot \delta_{z_i}, \,\;~~~~~ \hat{p}_i(s, a; \theta) = 
    \frac{\exp\left({l_i(s, a; \theta)}\right)}{\sum_{j=1}^m \exp \left({l_j(s, a; \theta)} \right)} \, .
\end{align*}%

To employ the cross-entropy loss (\autoref{eq:kl}) for TD learning, it is necessary that the target distribution is also a categorical distribution, supported on the same locations $z_i, \dots, z_m$. This allows for the direct computation of the cross-entropy loss as:
{\begin{align}
\boxed{\text{TD}_{\text{CE}}(\theta) = \mathbb{E}_{\mathcal{D}} \left[ \sum_{i=1}^{m} p_i(S_t, A_t; \theta^{-}) \log \hat{p}_i(S_t, A_t; \theta) \right]}\, ,     \label{q_param}
\end{align}}%
where the target probabilities $p_i$ are defined such that $\sum_{i=1}^m p_i(S_t, A_t; \theta^{-}) \,z_i \approx \tdtarget(S_t, A_t; \theta^{-})$.
In the subsequent sections, we explore two strategies for obtaining the target probabilities~$p_i(S_t, A_t; \theta^{-})$.

\begin{figure*}[t]
\centering
\includegraphics[width=0.95\linewidth]{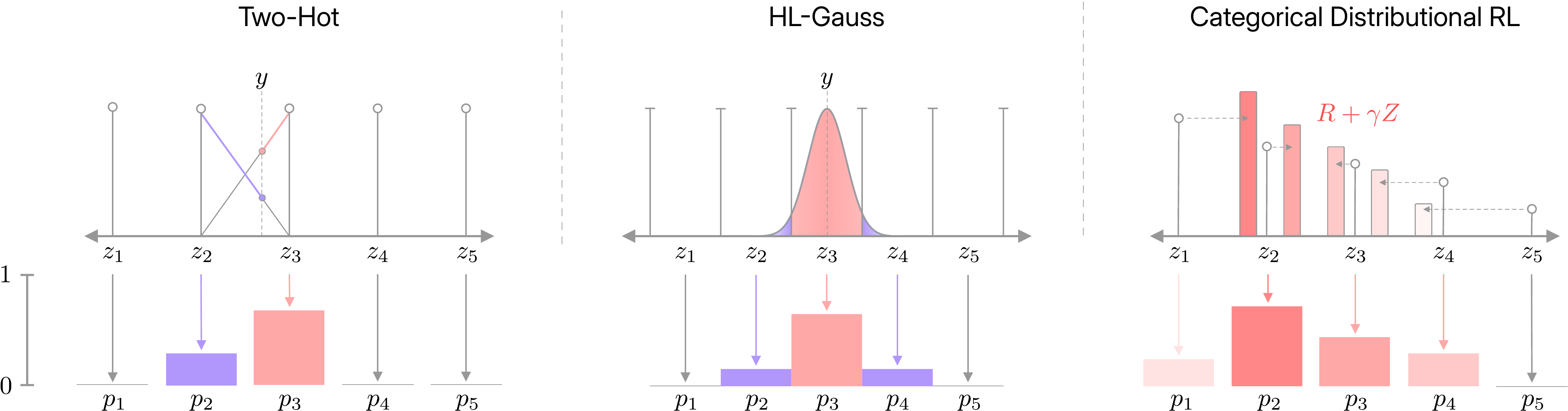}
\caption{\footnotesize{\textbf{Visualizing target-value categorical distribution in cross-entropy based TD learning}. While \twohot{} (left, \sref{subsec:transform_q}) puts probability mass on exactly two locations, \hlgauss{} (middle, \sref{subsec:transform_q}) distributes the probability mass to neighbouring locations (which is akin to smoothing the target value). \cdrl{} (right,  \sref{subsec:c51}) models the categorical return distribution, distributing probability mass proportionally to neighboring locations.}}
\vspace{-0.05cm}
\label{fig:methods}
\end{figure*}

\subsection{Constructing Categorical Distributions from Scalars}\label{subsec:transform_q}

The first set of methods we outline will project the scalar target $\tdtarget(S_t, A_t; \theta^{-})$ 
onto the categorical distribution supported on $\{ z_i \}_{i=1}^m$. A prevalent but na\"ive approach for the projection step involves discretizing the scalar into one of $m$ bins where $z_i$ represents the center of the bin. The resulting one-hot distribution is ``lossy'' and induces errors in the $Q$-function. These errors would compound as more Bellman backups are performed, resulting in more biased estimates, and likely worse performance.
To combat this, we first consider the ``two-hot'' approach~\citep{schrittwieser20muzero} that represents a scalar target \emph{exactly} via a unique categorical distribution that puts non-zero densities on two locations that the target lies between (see \autoref{fig:methods};~Left).

\noindent\textbf{A \twohot{} Categorical Distribution.} Let $z_i$ and $z_{i+1}$ be the locations which lower and upper-bound the TD target $z_i \le \tdtarget(S_t, A_t; \theta^{-}) \le z_{i+1}$. Then, the probability, $p_i$ and $p_{i+1}$, put on these locations is:
{\begin{align}\label{eqn:two_hot_probs}
    p_i(S_t, A_t; \theta^{-}) = \frac{\tdtarget(S_t, A_t; \theta^{-}) - z_i}{z_{i+1} - z_i}, \qquad
    p_{i+1}(S_t, A_t; \theta^{-}) = \frac{z_{i+1} - \tdtarget(S_t, A_t; \theta^{-})}{z_{i+1} - z_i}.
\end{align}}%
For all other locations, the probability prescribed by the categorical distribution is exactly zero. In principle, this \twohot{} transformation provides a uniquely identifiable and a non-lossy representation of the scalar TD target to a categorical distribution. However, \twohot{} does not fully harness the ordinal structure of discrete regression. Specifically, the classes are not independent and instead have a natural ordering, where each class intrinsically relates to its neighbors.

The class of Histogram Losses introduced by \citet{imani2018improving} seeks to exploit the ordinal structure of the regression task by distributing probability mass to neighboring bins -- akin to label smoothing in supervised classification \citep{szegedy16inception}. This is done by transforming a noisy version of the target value into a categorical distribution where probability mass can span multiple bins near the target (See \autoref{fig:methods}; Center), rather than being restricted to two locations.

\noindent\textbf{Histograms as Categorical Distributions.} Formally, define the random variable $Y \,|\, S_t, A_t$ with probability density $f_{Y|S_t, A_t}$ and cumulative distribution function $F_{Y|S_t, A_t}$ whose expectation is $\tdtarget(S_t, A_t; \theta^{-})$.
We can project the distribution $Y\,|\,S_t, A_t$ onto the histogram with bins of width $\varsigma = (v_{\text{max}} - v_{\text{min}})/m$ centered at $z_i$ by integrating over the interval $\left[ z_i - \nicefrac{\varsigma}{2}, z_i + \nicefrac{\varsigma}{2} \right]$ to obtain the probabilities,
{\begin{align}
p_i(S_t, A_t; \theta^{-}) &= \int_{z_i - \nicefrac{\varsigma}{2}}^{z_i + \nicefrac{\varsigma}{2}} f_{Y|S_t, A_t}(y\,|\,S_t, A_t) \,dy \tag*{} \\
&= F_{Y|S_t, A_t}(z_i + \nicefrac{\varsigma}{2} \,|\, S_t, A_t) - F_{Y|S_t, A_t}(z_i - \nicefrac{\varsigma}{2}\,|\,S_t, A_t) \, .
\end{align}}%

We now have a choice for the distribution $Y \,|\, S_t, A_t$. We follow the suggestion of \citet{imani2018improving} in using the Gaussian distribution $Y \,|\, S_t, A_t \sim \mathcal{N}(\mu = \tdtarget(S_t, A_t; \theta^{-}), \sigma^2)$ where the variance $\sigma^2$ is a hyper-parameter that can control the amount of label smoothing applied to the resulting categorical distribution. We refer to this method as \hlgauss{}.

\textbf{How should we tune $\sigma$ in practice?} \hlgauss{} requires tuning the standard deviation $\sigma$, in addition to the bin width $\varsigma$ and distribution range $[v_{min}, v_{max}]$. 99.7\% of the samples obtained by sampling from a standard Normal distribution should lie within three standard deviations of the mean with high confidence, which corresponds to approximately $6 \cdot \sigma/\varsigma$ bins. Thus, a more interpretable hyper-parameter that we recommend tuning is $\sigma/\varsigma$: setting it to $K/6$ distributes most of the probability mass to  $\ceil{K}+1$ neighbouring locations for a mean value centered at one of the bins. Unless specified otherwise, we set $\sigma/\varsigma = 0.75$ for our experiments, which distributes mass to approximately $6$ locations.

\vspace{-0.2cm}
\subsection{Modelling the Categorical Return Distribution}\label{subsec:c51}
\vspace{-0.2cm}

In the previous section, we chose to construct a target distribution from the usual scalar regression target representing the expected return. Another option is to directly model the distribution over future returns using our categorical model $\zapprox$, as done in distributional RL~\citep{ bellemare23distbook}. Notably, C51~\citep{bellemare17dist}, an early distributional RL approach, use the categorical representation along with minimizing the cross-entropy between the predicted distribution $\zapprox$ and the distributional analogue of the TD target.
To this end, we also investigate C51 as an alternative to \twohot{} and \hlgauss{} for constructing the target distribution for our cross-entropy objective.

\noindent\textbf{Categorical Distributional RL.} The first step to modelling the categorical return distribution is to define the analogous stochastic distributional Bellman operator on $\zapprox$,
{\begin{equation*}
    (\widehat{\mathcal{T}} \zapprox)(s, a; \theta^{-}) \stackrel{D}{=} \sum_{i=1}^m \hat{p}_i(S_{t+1}, A_{t+1}; \theta^{-}) \cdot \delta_{R_{t+1} + \gamma z_i } \,\,\big|\,\, S_t = s,\, A_t = a \,,
\end{equation*}}%
where $A_{t+1} = \argmax_{a'}\qapprox(S_{t+1}, a')$.
As we can see, the stochastic distributional Bellman operator has the effect of shifting and scaling the locations $z_i$ necessitating the categorical projection, first introduced by \citet{bellemare17dist}.
At a high level, this projection distributes probabilities proportionally to the immediate neighboring locations $z_{j-1} \le R_{t+1} + \gamma z_i \le z_j$ (See \autoref{fig:methods}; Right). To help us identify these neighboring locations we define $\floor{x} = \argmax\{z_i : z_i \le x\}$ and $\ceil{x} = \argmin\{z_i : z_i \ge x \}$.
Now the probabilities for location $z_i$ can be written as,
{\begin{align}\label{eqn:c51_probs}
    &p_{i}(S_t, A_t; \theta^{-}) = \sum_{j=1}^m \hat{p}_j(S_{t+1}, A_{t+1}; \theta^{-}) \cdot \xi_{j}(R_{t+1} + \gamma z_i) \\
    &\xi_{j}(x) = \frac{x - z_j}{z_{j+1} - z_j} \ind\{ \floor{x} = z_j \}
    + \frac{z_{j+1} - x}{z_{j+1} - z_j}\ind\{ \ceil{x} = z_{j} \} \, . \tag*{}
\end{align}}%
For a complete exposition of the categorical projection, see \citet[][Chapter 5]{bellemare23distbook}.

\vspace{-0.2cm}
\section{Evaluating Classification Losses in RL} \label{experiments}
\vspace{-0.2cm}

The goal of our experiments in this section is to evaluate the efficacy of the various target distributions discussed in Section~\ref{sec:td_target} combined with the categorical cross-entropy loss \eqref{q_param} in improving performance and scalability of value-based deep RL on a variety of problems. This includes several single-task and multi-task RL problems on Atari 2600 games as well as domains beyond Atari including language agents, chess, and robotic manipulation. These tasks consist of both online and offline RL problems. For each task, we instantiate our cross-entropy losses in conjunction with a strong value-based RL approach previously evaluated on that task. 
Full experimental methodologies including hyperparameters for each domain we consider can be found in Appendix~\ref{app:exp}.

\vspace{-0.5cm}
\subsection{Single-Task RL on Atari Games}
\label{subsec:single_task_atari}
\vspace{-0.2cm}
\begin{figure*}[t]
\centering
\begin{minipage}{.55\textwidth}
    \centering
    \vspace{-0.2cm}
    \includegraphics[width=\textwidth]{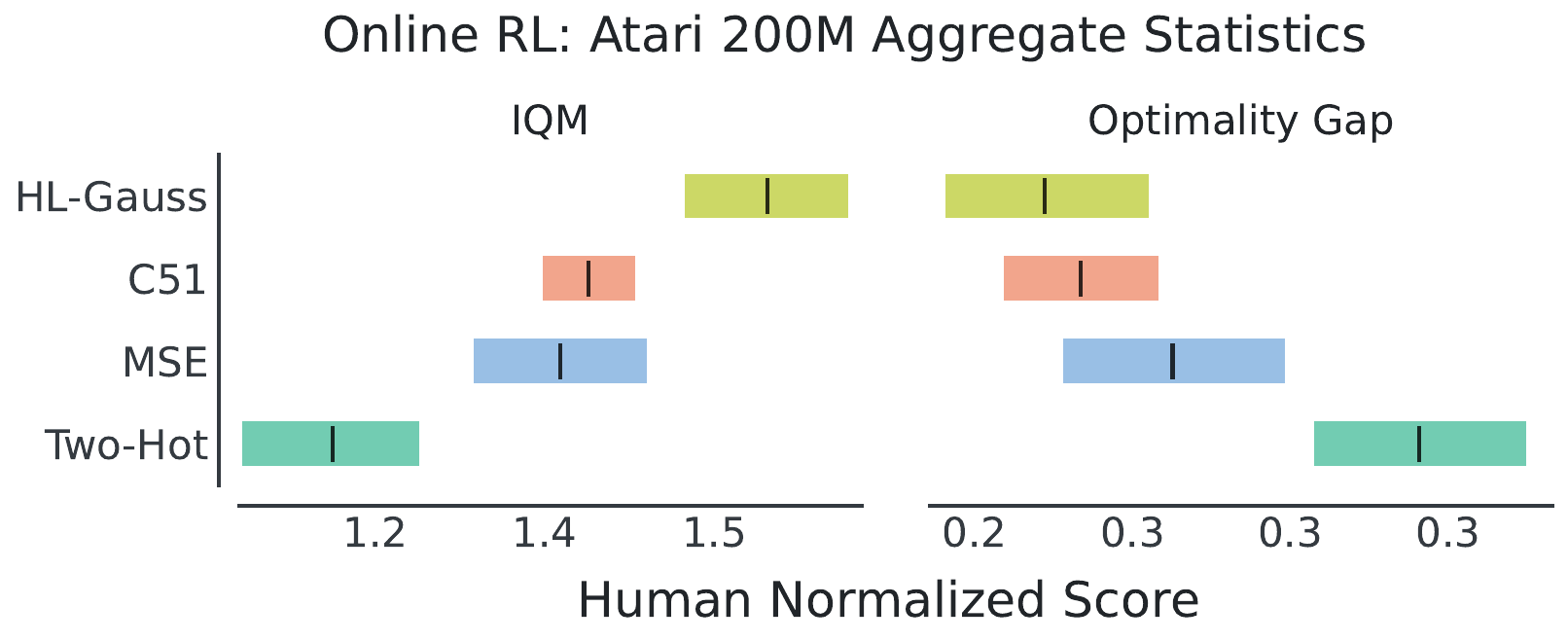}
\end{minipage}%
\hfill%
\begin{minipage}{.44\textwidth}
    \centering
    \includegraphics[width=\textwidth]{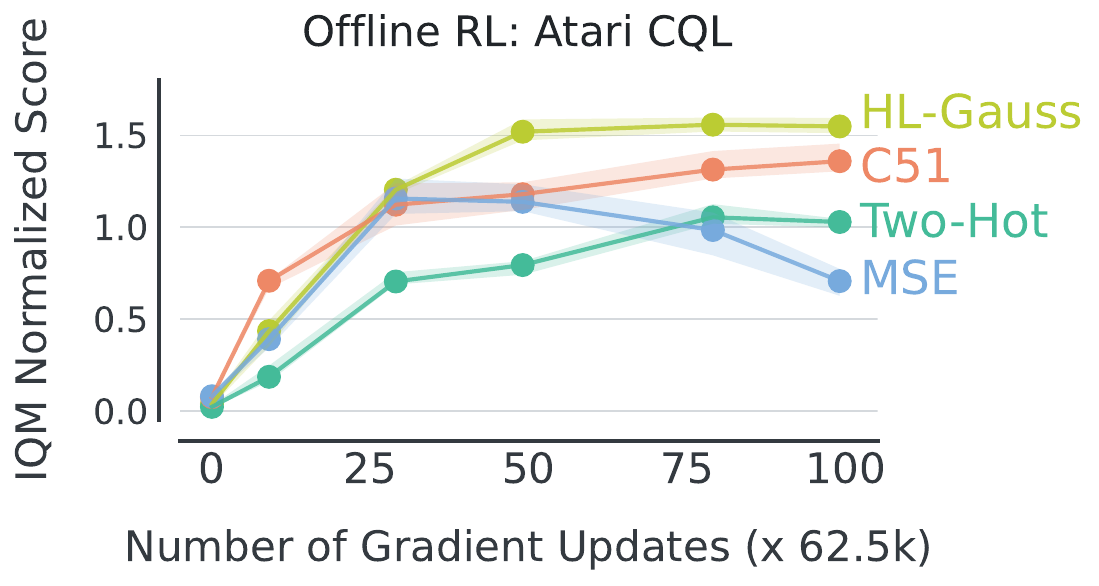}
\end{minipage}%
\vspace{-0.2cm}
\caption{\footnotesize{\textbf{Regression vs cross-entropy losses for} \textbf{\textbf{(Left)} Online RL and \textbf{(Right)} Offline RL~(\sref{subsec:single_task_atari})}. \hlgauss{} and \cdrl{} outperform MSE, with \hlgauss{} performing the best. Moreover, \twohot{} loss underperforms MSE but is more stable with prolonged training in offline RL, akin to other cross-entropy losses.
See \sref{subsec:single_task_atari} for more details.
}}
\label{fig:atari_main}
\vspace{-0.45cm}
\end{figure*}

We first evaluate the efficacy of \hlgauss{}, \twohot{}, and C51~\citep{bellemare17dist} -- an instantiation of categorical distributional RL, on the Arcade Learning Environment~\citep{bellemare2013arcade}.
For our regression baseline we train DQN \citep{mnih15dqn} on the mean-squared error TD objective which has been shown to outperform other regression based losses \citep{ceron2021revisiting}.
Each method is trained with the Adam optimizer, which has been shown to reduce the performance discrepancy between regression-based methods and distributional RL approaches~\citep{agarwal2021precipice}.

\textbf{Evaluation}. Following the recommendations by \citet{agarwal2021precipice}, we report the interquartile mean~(IQM) normalized scores with 95\% stratified bootstrap confidence intervals (CIs), aggregated across games with multiple seeds each. We report human-normalized aggregated scores across 60 Atari games for online RL. For offline RL, we report behavior-policy normalized scores aggregated across 17 games, following the protocol in \citet{kumar2021implicit}.

\textbf{Online RL results}. Following the setup of \citet{mnih15dqn}, we train DQN for 200M frames with the aforementioned losses.
We report aggregated human-normalized IQM performance and optimality gap %
across 60 Atari games in Figure~\ref{fig:atari_main}.  Observe that \hlgauss{} substantially outperforms the \twohot{} and MSE losses. Interestingly, \hlgauss{} also improves upon categorical distributional RL (C51), despite not modelling the return distribution. This finding suggests that the loss (categorical cross-entropy) is perhaps the more crucial factor for C51, as compared to modelling the return distribution.  

\textbf{Offline RL results.} The strong performance of \hlgauss{} with online DQN, which involves learning from self-collected interactions, raises the question of whether it would also be effective in learning from offline datasets. To do so, we train agents with different losses on the 10\% Atari DQN replay dataset~\citep{agarwal2020optimistic} using CQL~(\sref{sec:rl_back}) for 6.25M gradient steps. 
As shown in \autoref{fig:atari_main}, \hlgauss{} and C51 consistently outperform MSE, while \twohot{} shows improved stability over MSE but underperforms other classification methods. Notably, \hlgauss{} again surpasses C51 in this setting. Furthermore, consistent with the findings of \citet{kumar2021implicit}, utilizing the mean squared regression loss results in performance degradation with prolonged training. However, cross-entropy losses (both \hlgauss{} and C51) do not show such degradation and generally, remain stable.

\vspace{-0.2cm}
\subsection{Scaling Value-based RL to Large Networks}
\label{sec:generalist}
\vspace{-0.2cm}

In supervised learning, particularly for language modeling~\citep{kaplan2020scaling}, increasing the parameter count of a network typically improves performance. However, such scaling behavior remain elusive for value-based deep RL methods, where \emph{naive} parameter scaling can hurt
performance~\citep{taiga2023multitask, kumar2022offline, obando2024mixtures}. To this end, we investigate the efficacy of our classification methods, as an alternative to MSE regression loss in deep RL, towards enabling better performance with parameter scaling for value-networks.

\subsubsection{Scaling with Mixture-of-Experts}
\label{moe}

\begin{figure*}[t]
    \centering
    \begin{minipage}[b]{.49\textwidth}
    \centering
    \includegraphics[width=0.98\linewidth]{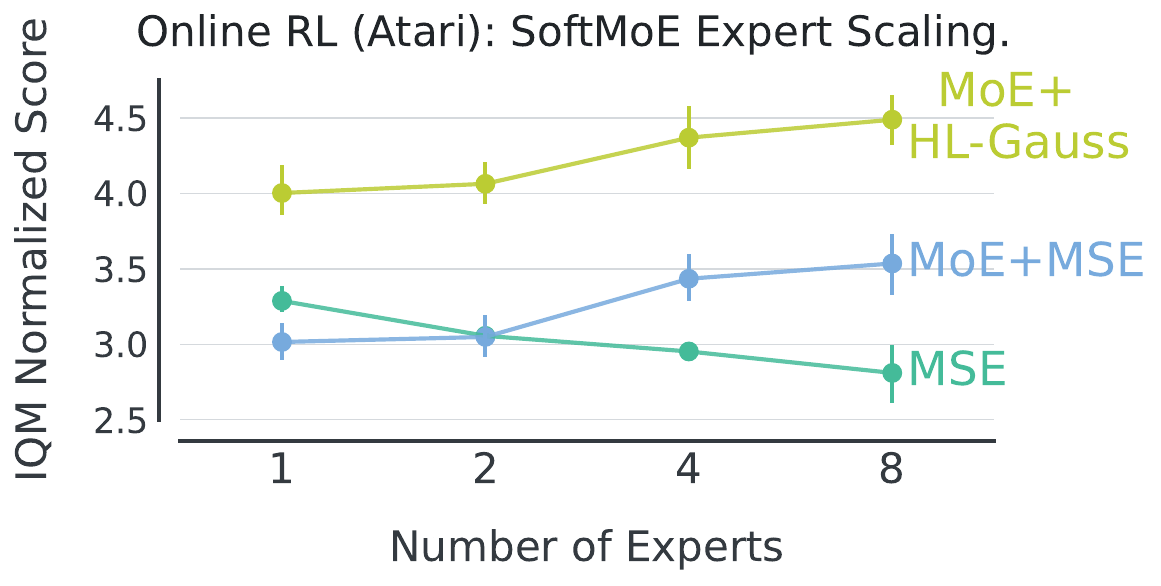}
    \vspace{0.65mm}
    \caption{\footnotesize{\textbf{MoE scaling curves for \hlgauss{} and MSE on Online RL}. \hlgauss{}, with a single expert, outperform all regression configurations. Both \hlgauss{} and MSE scale similarly when employing SoftMoE, with \hlgauss{} providing $\approx30\%$ IQM improvement. SoftMoE also mitigates negative scaling observed with MSE alone. See \sref{moe} for more details.}}
    \label{fig:moe}
    \end{minipage}%
    \hfill%
    \begin{minipage}[b]{0.49\textwidth}
    \centering
    \includegraphics[width=0.98\linewidth]{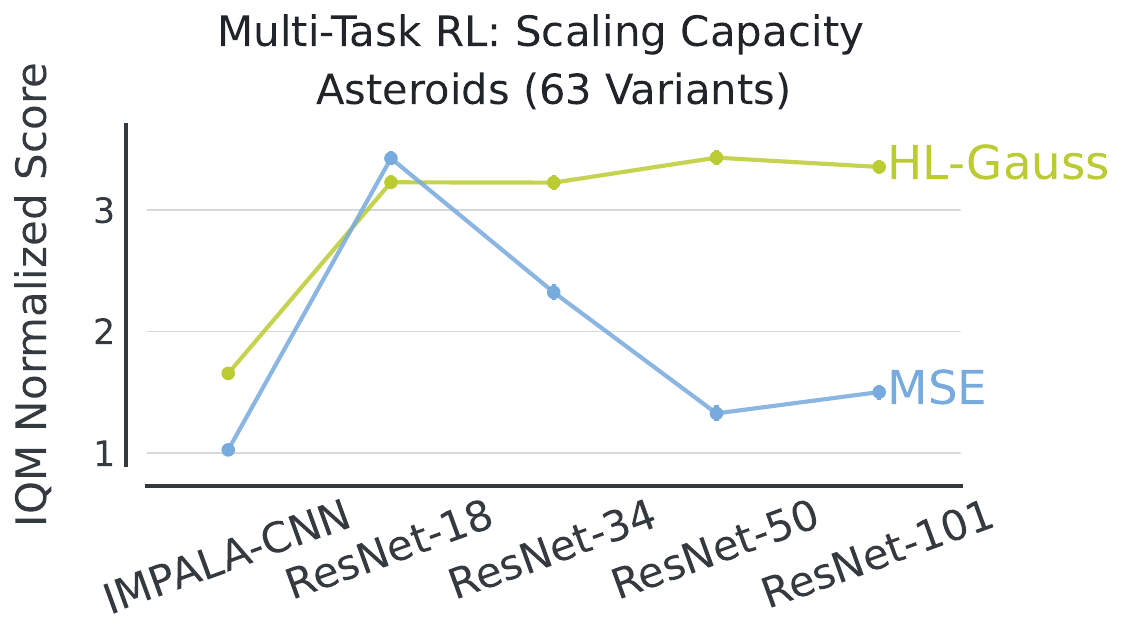}
    \caption{\footnotesize{\textbf{Scaling curves on Multi-task Online RL}. %
    Results for actor-critic IMPALA with ResNets on \textsc{Asteroids}. HL-Gauss outperforms MSE and notably reliably scales better with larger networks. Since human scores are not available for  variants, we report normalized scores using a baseline IMPALA agent with MSE loss. See \sref{subsubsec:multi-task} for more details.}} 
    \label{fig:online_scaling}
    \end{minipage}
\end{figure*}

Recently, \citet{obando2024mixtures} demonstrate that while parameter scaling with convolutional networks hurts single-task RL performance on Atari, incorporating Mixture-of-Expert~(MoE) modules in such networks improves performance. Following their setup, we replace the penultimate layer in the architecture employed by Impala \citep{espeholt2018impala} with a SoftMoE~\citep{puigcerver2023sparse} module and vary the number of experts in $\{1, 2, 4, 8\}$. Since each expert is a copy of the original penultimate layer, this layer's parameter count increases by a factor equal to the number of experts.
The only change we make is to replace the MSE loss in SoftMoE DQN, as employed by \citet{obando2024mixtures}, with the \hlgauss{} cross-entropy loss.
We train on the same subset of 20 Atari games used by \citet{obando2024mixtures} and report aggregate results over five seeds~in~\autoref{fig:moe}.

As shown in \autoref{fig:moe}, we find that \hlgauss{} consistently improves performance over MSE by a constant factor independent of the number of experts. One can also observe that SoftMoE + MSE seems to mitigate some of the negative scaling effects observed with MSE alone. As SoftMoE + MSE uses a softmax in the penultimate layer this could be providing similar benefits to using a classification loss but as we will later see these benefits alone cannot be explained by the addition of the softmax.

\subsubsection{Training Generalist Policies with ResNets}
\label{subsubsec:multi-task}
Next, we consider scaling value-based ResNets~\citep{resnet} in both offline and online settings to train a generalist video game-playing policy on Atari. In each case, we train a family of differently sized Q-networks for multi-task RL, and report performance as a function of the network size. 

\textbf{Multi-task Online RL}. Following \citet{taiga2023multitask}, we train a multi-task policy capable of playing Atari game variants with different environment dynamics and rewards \citep{farebrother18generalization}. We evaluate two Atari games: 63 variants for \textsc{Asteroids} and 29 variants for \textsc{Space Invaders}. We employ a distributed actor-critic method, IMPALA \citep{espeholt2018impala}, and compare the standard MSE critic loss with the cross-entropy based \hlgauss{} loss. Our experiments investigate the scaling properties of these losses when moving from Impala-CNN ($\le$ 2M parameters) to larger ResNets~\citep{resnet} up to ResNet-101~(44M parameters).
We evaluate multi-task performance after training for 15 billion frames, and repeat each experiment with five seeds.

Results for \textsc{Asteroids} are presented in \autoref{fig:online_scaling}, with additional results on \textsc{Space Invaders} presented in \autoref{fig:online_scaling_with_spaceinvaders}. We observe that in both environments \hlgauss{} consistently outperforms MSE.
Notably, \hlgauss{} scales better, especially on \textsc{Asteroids} where it even slightly improves performance with larger networks beyond ResNet-18, while MSE performance significantly degrades.

\textbf{Multi-game Offline RL}. We consider the the setup from \citet{kumar2022offline}, where we modify their recipe to use a non-distributional \hlgauss{} loss, in place of distributional C51. Specifically, we train a single generalist policy to play 40 different Atari games simultaneously, when learning from a ``near-optimal'' training dataset, composed of replay buffers obtained from online RL agents trained independently on each game. This multi-game RL setup was originally proposed by \citet{lee2022multigame}. The remaining design choices (e.g., feature normalization; the size of the network) are kept identical.

\begin{figure*}[h]
    \centering
    \begin{minipage}[c]{0.375\linewidth}
    \centering
    \includegraphics[width=0.95\textwidth]{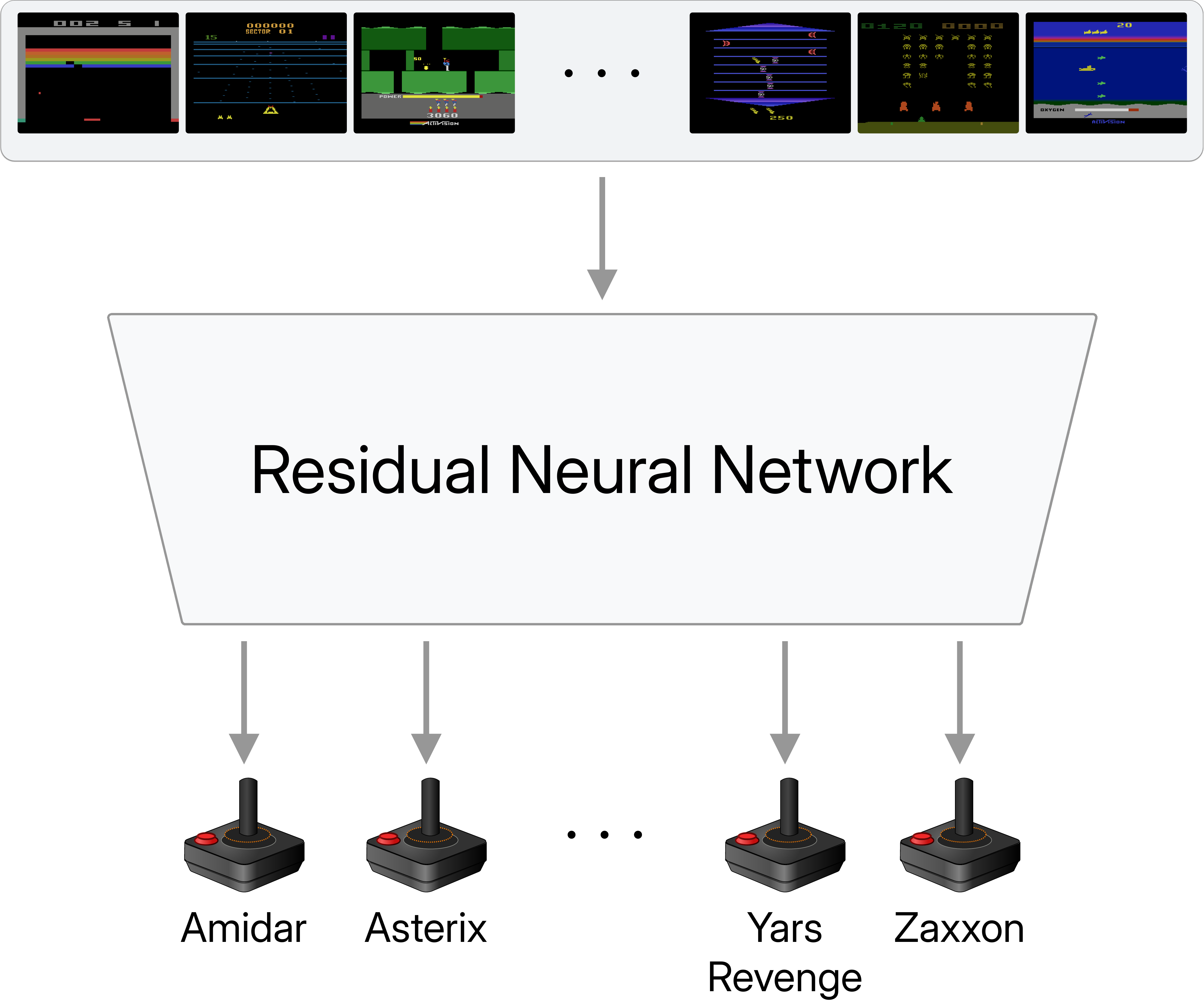}
    \end{minipage}
    \hfill
    \begin{minipage}[c]{0.575\linewidth}
    \centering
    \includegraphics[width=\textwidth]{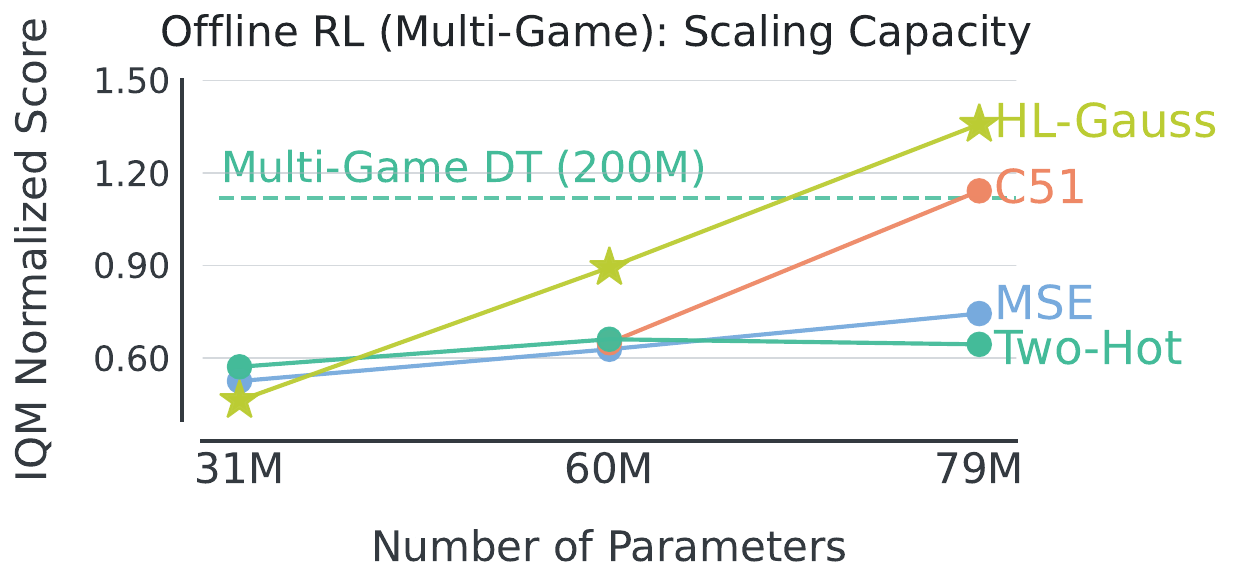}
    \end{minipage}
\caption{\footnotesize{\textbf{Scaling curves on Multi-game Atari (Offline RL)}. IQM human normalized score for ResNet-$\{34,50,101\}$, with spatial embeddings, to play 40 Atari games simultaneously using a single value network~\citep{kumar2022offline}. \hlgauss{} enables remarkable scaling, substantially outperforming categorical distributional RL (C51) and regression (MSE) losses used by prior work, as well as the multi-game Decision Transformer~\citep{lee2022multigame}. See \sref{subsubsec:multi-task} for more details and Figure~\ref{fig:offline_scaling_dqn_norm} for a version of these results reported in terms of DQN normalized scores, another commonly used metric.}}
\label{fig:offline_scaling}
\end{figure*}

As shown in Figure~\ref{fig:offline_scaling}, \hlgauss{} scales even better than the C51 results from \citet{kumar2022offline}, resulting in an improvement of about $45\%$ over the best prior multi-game result available with ResNet-101 (80M parameters) as measured by the IQM human normalized score~\citep{agarwal2021precipice}. Furthermore, while the performance of MSE regression losses typically plateaus upon increasing model capacity beyond ResNet-34, \hlgauss{} is able to leverage this capacity to improve performance, indicating the efficacy of classification-based cross-entropy losses. Additionally, when normalizing against scores obtained by a DQN agent, we show in Figure~\ref{fig:offline_scaling_dqn_norm} that in addition to performance, the rate of improvement as the model scale increases tends to also be larger for the \hlgauss{} loss compared to C51. 

\subsection{Value-Based RL with Transformers}
\label{sec:beyond_atari}

Next, we evaluate the applicability of the \hlgauss{} cross-entropy loss beyond Atari. To do so, we consider several tasks that utilize high-capacity Transformers, namely, a language-agent task of playing Wordle, playing Chess without inference-time search, and robotic manipulation.

\subsubsection{Language Agent: Wordle}
\label{subsubsec:wordle}
\begin{figure*}[t]
    \centering
    \begin{minipage}[c]{0.4\textwidth}
        \centering
        \includegraphics[width=0.9\linewidth]{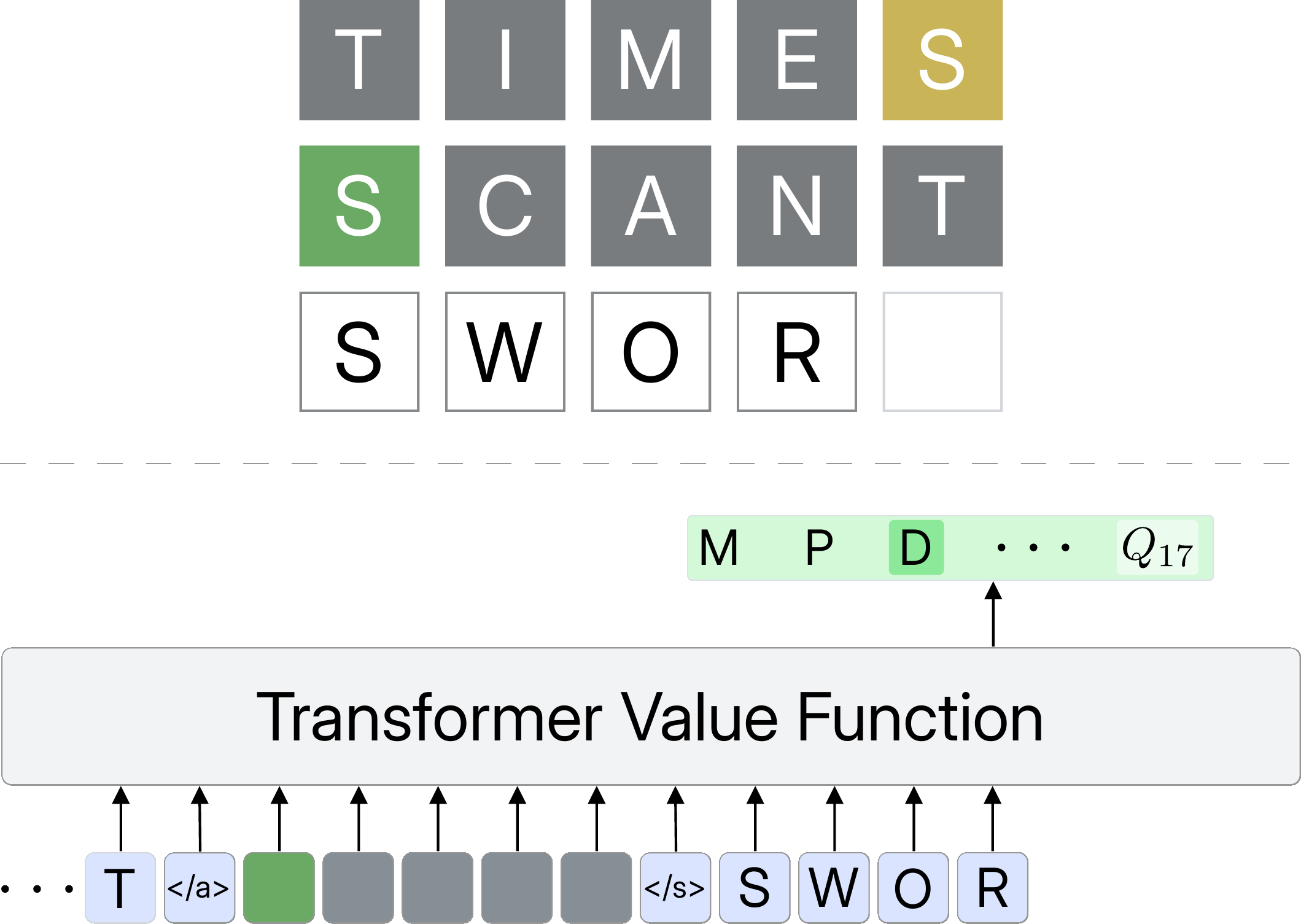}
    \end{minipage}
    \hfill
    \begin{minipage}[c]{.575\textwidth}
        \centering
    \includegraphics[width=1.0\linewidth]{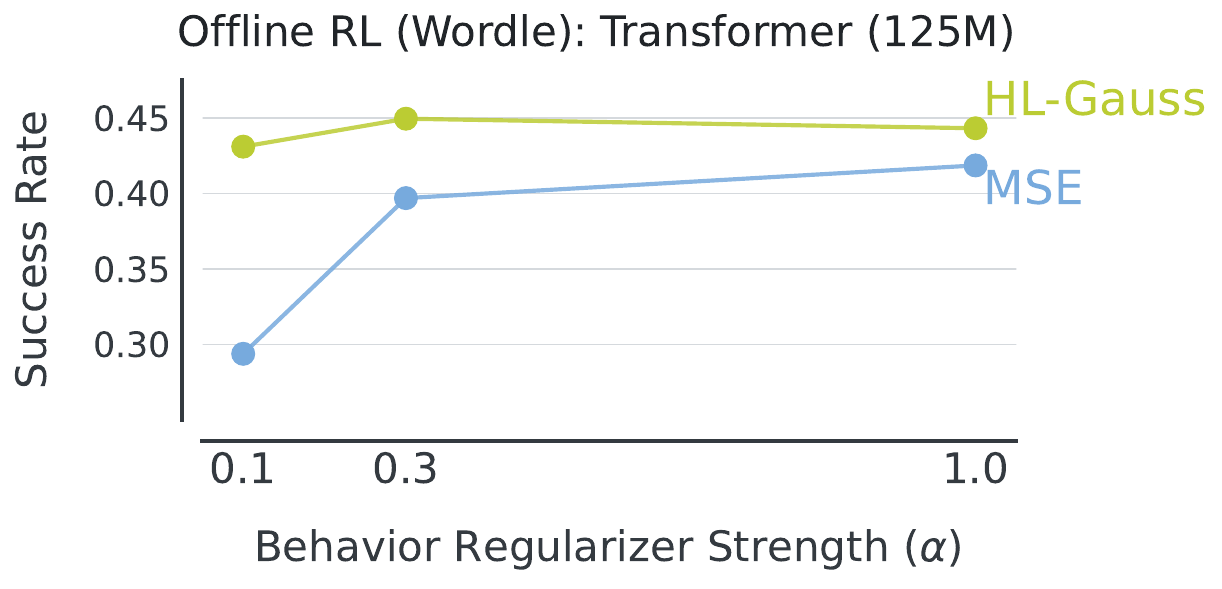}
    \end{minipage}%
    \caption{\footnotesize{\textbf{Regression vs cross-entropy loss for Wordle (Offline RL).} Comparing \hlgauss{} cross-entropy loss with MSE regression loss for a transformer trained with offline RL on Wordle dataset~\citep{snell2022offline}. Here, we evaluate the success rate of guessing the word in one turn given a partially played Wordle game (e.g., image on left). \hlgauss{} leads to substantially higher success rates for varying strengths of behavior regularization. See \sref{subsubsec:wordle} for more details.}}
    \label{fig:wordle}
\end{figure*}

To evaluate whether classification losses enhance the performance of value-based RL approaches on language agent benchmarks, we compare \hlgauss{} with MSE on the task of playing the game of Wordle\footnote{\url{www.nytimes.com/games/wordle/index.html}}.  Wordle is a word guessing game in which the agent gets 6 attempts to guess a word. Each turn the agent receives environment feedback about whether guessed letters are in the true word. The dynamics of this task are non-deterministic. More generally, the task follows a turn-based structure, reminiscent of dialogue tasks in natural language processing. This experiment is situated in the offline RL setting, where we utilize the dataset of suboptimal game-plays provided by \citet{snell2022offline}. Our goal is to train a GPT-like, decoder-only Transformer, with 125M parameters, representing the Q-network. 
See Figure~\ref{fig:wordle}~(left) for how the transformer model is used for playing this game. %

On this task, we train the language-based transformer for 20K gradient steps with an offline RL approach combining Q-learning updates from DQN with a CQL-style behavior regularizer~(\sref{sec:rl_back}), which corresponds to standard next-token prediction loss (in this particular problem).  As shown in Figure~\ref{fig:wordle}, \hlgauss{} outperforms MSE, for multiple coefficients controlling the strength of CQL regularization.

\subsubsection{Grandmaster-level Chess without Search}
\label{subsubsec:chess}

\begin{figure*}
    \centering
    \begin{minipage}[c]{.4\textwidth}
    \centering
    \includegraphics[width=\linewidth]{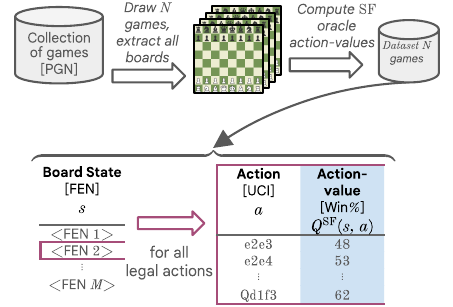}
    \end{minipage}
    \hfill
    \begin{minipage}[c]{0.575\textwidth}
        \centering
        \includegraphics[width=\linewidth]{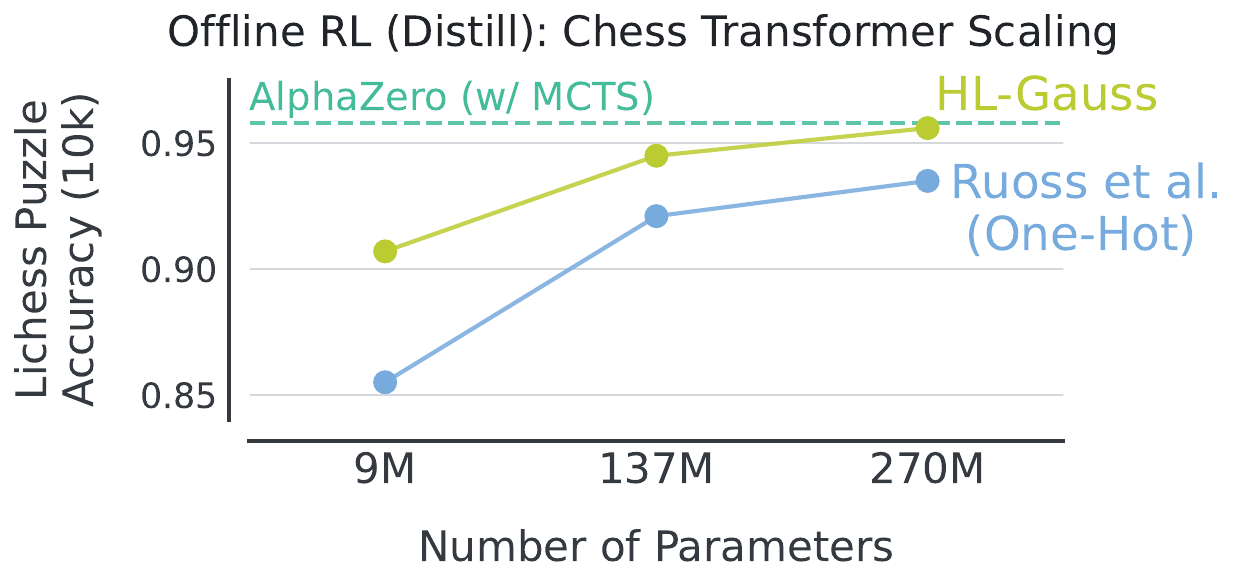}
    \end{minipage}
    \caption{\footnotesize{\textbf{Grandmaster-level Chess without Search.} \textbf{(Left)}  Dataset generation for Q-value distillation on Chess. \textbf{(Right)} Scaling Curves. Following the setup from \citet{ruoss2024grandmaster}, where they train Transformer models to play chess via supervised learning on Stockfish 16 Q-values and then follow greedy policy for evaluation. As the results show, \hlgauss{} outperforms one-hot targets used by \citet{ruoss2024grandmaster} and nearly matches the performance of AlphaZero with tree search. }}
    \label{fig:chess}
\end{figure*}

Transformers have demonstrated their effectiveness as general-purpose algorithm approximators, effectively amortizing expensive inference-time computation through distillation~\citep{ruoss2024grandmaster, lehnert2024beyond}.
In this context, we explore the potential benefits of using \hlgauss{} to convert scalar action-values into classification targets for distilling a value-function.
Using the setup of \citet{ruoss2024grandmaster}, we evaluate \hlgauss{} for distilling the action-value function of Stockfish 16 --- the strongest available Chess engine that uses a combination of complex heuristics and explicit search --- into a causal transformer.
The distillation dataset comprises 10 million chess games annotated by the Stockfish engine, yielding 15 billion data points~(\autoref{fig:chess}, left).

We train 3 transformer models of varying capacity (9M, 137M, and 270M parameters) on this dataset, using either \hlgauss{} or 1-Hot classification targets. We omit MSE as \citet{ruoss2024grandmaster} demonstrate that 1-Hot targets outperform MSE on this task.
The effectiveness of each model is evaluated based on its ability to solve 10,000 chess puzzles from Lichess, with success measured by the accuracy of the generated action sequences compared to known solutions. Both the setup and results are presented in \autoref{fig:chess} (right).
While the one-hot target with the 270M Transformer from \citet{ruoss2024grandmaster} outperformed an AlphaZero baseline without search, \hlgauss{} closes the performance gap with the substantially stronger AlphaZero with 400 MCTS simulations~\citep{schrittwieser20muzero}.

\subsubsection{Generalist Robotic Manipulation with Offline Data}
\label{subsubsec:robot}
Finally, we evaluate whether cross-entropy losses can improve performance on a set of large-scale vision-based robotic manipulation control tasks from \citet{chebotar2023q}. These tasks present a simulated 7-DoF mobile manipulator, placed in front of a countertop surface. The goal is to control this manipulator to successfully grasp and lift 17 different kitchen objects in the presence of distractor objects, clutter, and randomized initial poses.
We generate a dataset of $500,000$ (successful and failed) episodes starting from a small amount of human-teleoperated demonstrations ($40,000$ episodes) by replaying expert demonstrations with added sampled action noise, reminiscent of failed autonomously-collected rollouts obtained during deployment or evaluations of a behavioral cloning policy trained on the human demonstration data.

\begin{figure*}
    \centering
    \begin{minipage}[c]{.4\textwidth}
    \centering
    \includegraphics[width=0.8\linewidth]{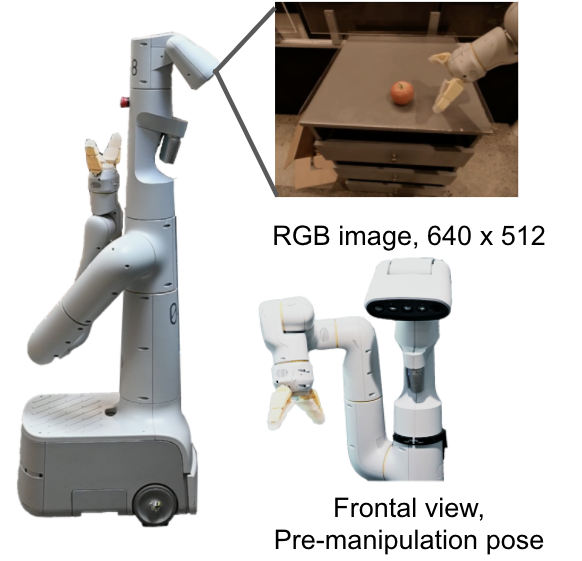}
    \end{minipage}%
    \hfill
    \begin{minipage}[c]{0.575\textwidth}
        \centering
        \includegraphics[width=\linewidth]{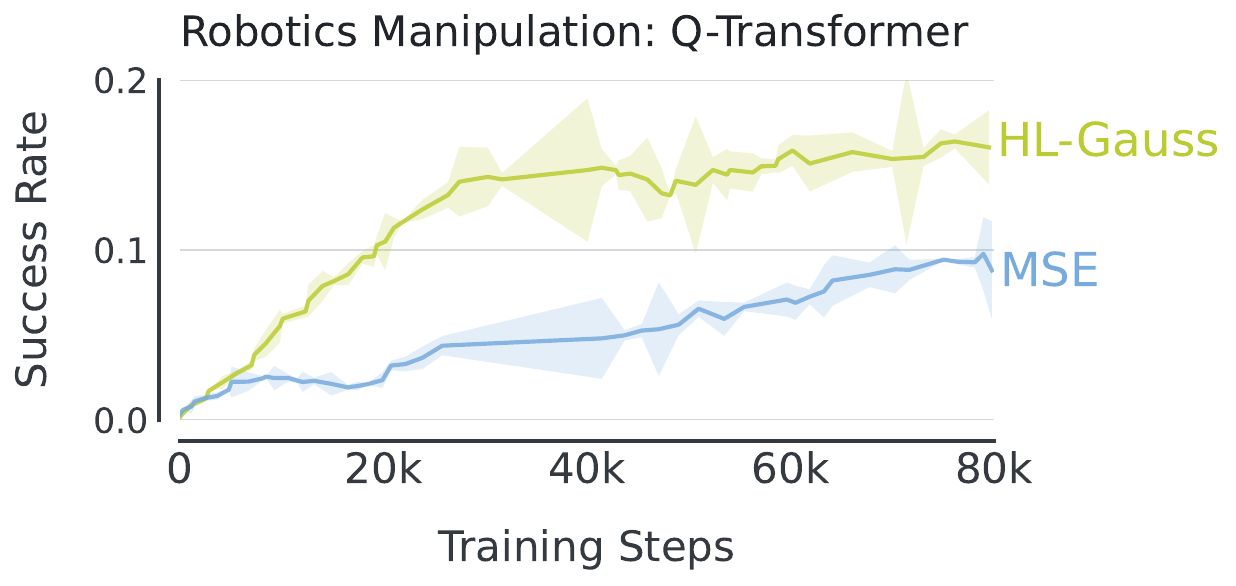}
    \end{minipage}
\caption{\footnotesize{\textbf{Generalist robotic manipulation with offline data: \textbf{(Left)} Robot platform and \textbf{(Right)} \hlgauss{} vs MSE on simulated vision-based manipulation.} The robotic manipulation problem~(\sref{subsubsec:robot}) uses the setup from \citet{chebotar2023q}. The image on the left shows the 7 degree of freedom mobile manipulator robot used for these experiments. In the plots, error bars show 95\% CIs. Note that utilizing a \hlgauss{} enables significantly faster learning to a better point.}}
        \label{fig:q_transformer}
\end{figure*}

We train a Q-Transformer model with 60M parameters, following the recipe in \citet{chebotar2023q}, but replace the MSE regression loss with the \hlgauss{} classification loss. As shown in \autoref{fig:q_transformer}, \hlgauss{} results in $67\%$ higher peak performance over the regression baseline, while being much more sample-efficient, addressing a key limitation of the prior regression-based approach.

\vspace{-0.2cm}
\section{Why Does Classification Benefit RL?}
\vspace{-0.2cm}
Our experiments demonstrate that classification losses can significantly improve the performance and scalability of value-based deep RL. In this section, we perform controlled experiments to understand why classification benefits value-based RL. Specifically, we attempt to understand how the categorical cross-entropy loss can address several challenges specific to value-based RL including representation learning, stability, and robustness. We will also perform ablation experiments to uncover the reasons behind the superiority of \hlgauss{} over other categorical targets.

\subsection{Ablation Study: What Components of Classification Losses Matter?}

Classification losses presented in this paper differ from traditional regression losses used in value-based RL in two ways: \textbf{(1)} parameterizing the output of the value-network to be a categorical distribution in place of a scalar, and \textbf{(2)} strategies for converting scalar targets into a categorical target. We will now understand the relative contribution of these steps towards the performance of cross-entropy losses. 

\subsubsection{Are Categorical Representations More Performant?}
\label{sec:softmax_stability}

\begin{figure*}[t]
    \begin{minipage}[b]{.48\textwidth}
        \centering
        \includegraphics[width=\linewidth]{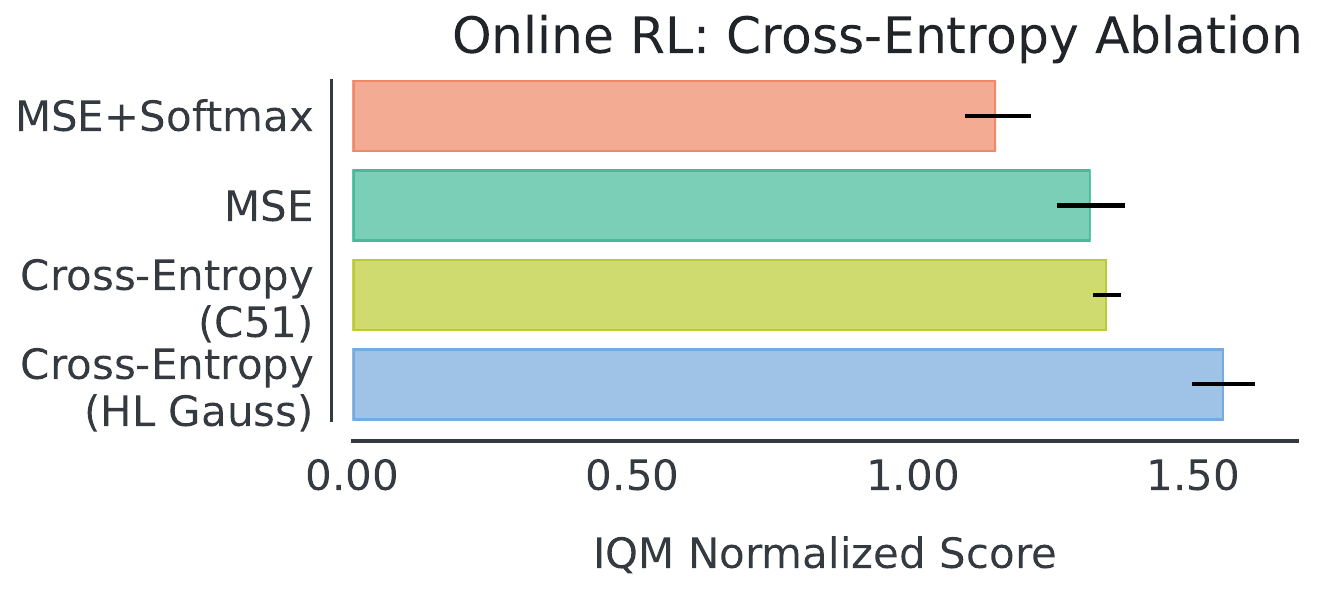}
        \vspace{-0.5cm}
        \caption{\footnotesize{\textbf{Evaluating the learning stability of softmax parameterization ~(\sref{sec:softmax_stability}) in online RL on Atari.} Categorical representation of Q-values does not benefit MSE + Softmax relative to MSE, implying that the cross-entropy loss is critical.}}
        \label{fig:softmax}
    \end{minipage}\hfill
    \begin{minipage}[b]{0.48\textwidth}
        \centering
        \includegraphics[width=\linewidth]{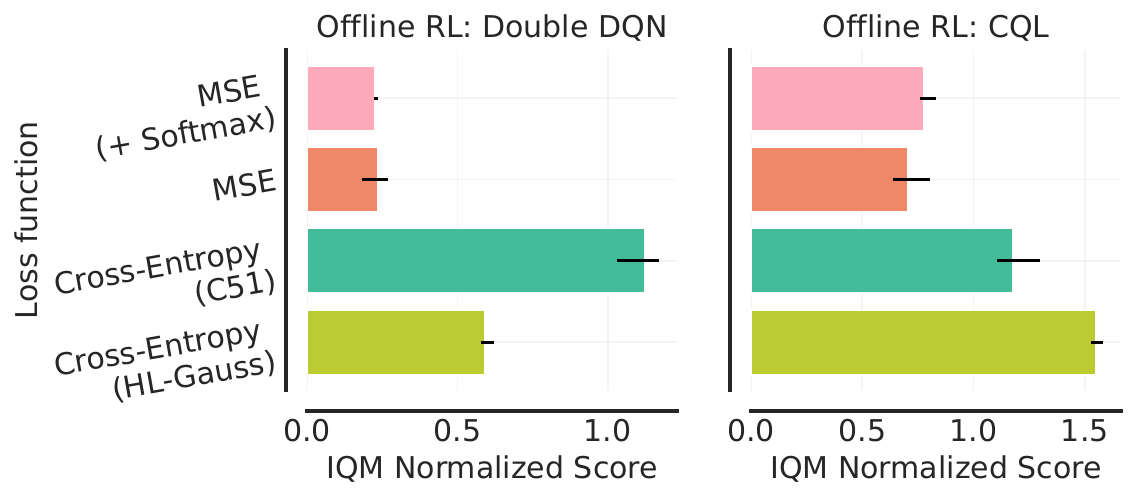}
        \caption{\footnotesize{\textbf{Evaluations of the learning stability of MSE+Softmax~(\sref{sec:softmax_stability}) in Offline RL on Atari.} We do not observe any substantial gains from using a softmax operator with the MSE loss for either architecture. This implies that the cross-entropy loss is critical.}}
        \label{fig:softmax_app}
    \end{minipage}
\end{figure*}

As discussed in \sref{subsec:transform_q}, we parameterize the Q-network to output logits that are converted to probabilities of a categorical distribution by applying the ``softmax'' operator. Using softmax leads to bounded Q-values and bounded output gradients, which can possibly improve RL training stability~\citep{ hansen23tdmpc2}.
To investigate whether our Q-value parameterization alone results in improved performance without needing a cross-entropy loss, we train Q-functions with the same parameterization as Eq~(\ref{q_param}) but with MSE.  We do not observe any gains from using softmax in conjunction with the MSE loss in both online~(\autoref{fig:softmax}) and offline RL~(\autoref{fig:softmax_app}). This highlights that the use of the cross-entropy loss results in the bulk of the performance improvements. 

\begin{figure*}[t]
    \centering
    \includegraphics[width=0.65\linewidth]{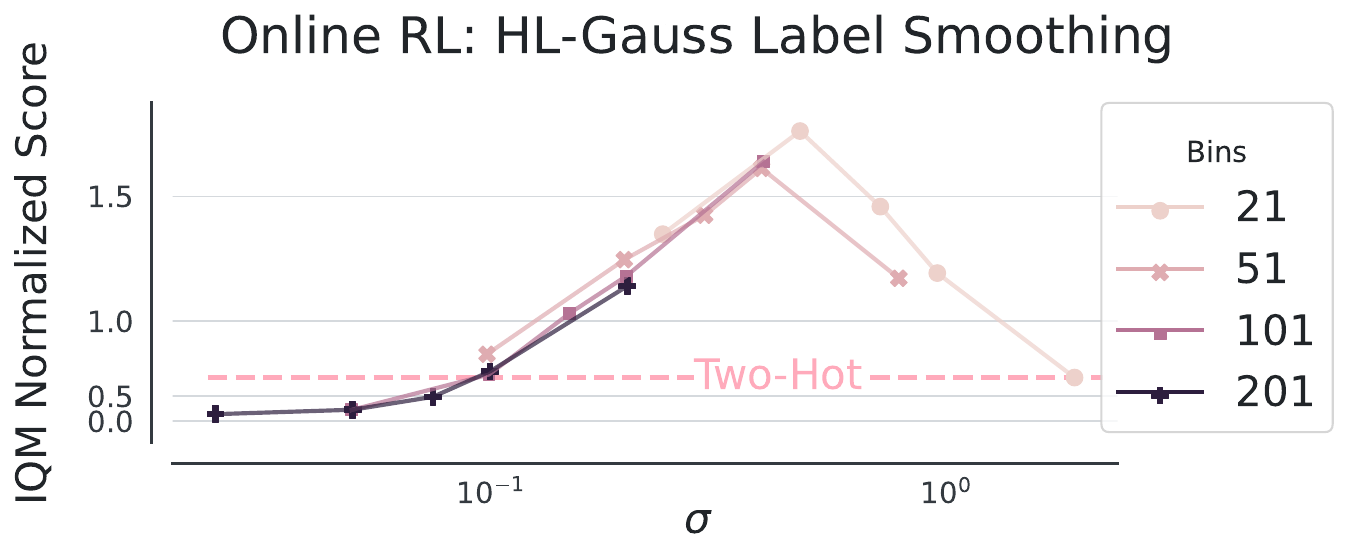}
    \vspace{-0.55cm}
    \caption{\footnotesize{\textbf{Sweeping the ratio $\sigma / \varsigma$ for different number of bins in Online RL on Atari.}. A wide range of $\sigma$ values outperform \twohot{}, which corresponds to not using any label smoothing, implying that \hlgauss{} does benefit from a label smoothing like effect. Furthermore, the optimal amount of label smoothing as prescribed by $\sigma$ is independent of bin width $\varsigma$. This implies that the \hlgauss{} is leveraging the structure of the regression problem and the gains cannot be purely attributed to reduced overfitting from label smoothing~(\sref{sec:distributed_targets}).}}
    \label{fig:hl_label_smooth}
\end{figure*}

\subsubsection{Why Do Some Cross-Entropy Losses Work Better Than Others?}
\label{sec:distributed_targets}

Our results indicate that \hlgauss{} outperforms \twohot{}, despite both these methods using a cross-entropy loss. We hypothesize that the benefits of \hlgauss{} could stem from two reasons: 1) \hlgauss{} reduces overfitting by spreading probability mass to neighboring locations; and 2) \hlgauss{} generalizes across a specific range of target values, exploiting ordinal structure in the regression problem. The first hypothesis would be more consistent with how label smoothing addresses overfitting in classification problems~\citep{szegedy16inception}. 

We test these hypotheses in the online RL setting across a subset of $13$ Atari games. To do so, we fix the value range $[v_{\text{min}}, v_{\text{max}}]$ while simultaneously varying the number of categorical bins in $\{ 21, 51, 101, 201 \}$ and the ratio of deviation $\sigma$ to bin width $\varsigma$ in $\{ 0.25, 0.5, 0.75, 1.0, 2.0 \}$. 
We find that a wide range of $\sigma$ values for \hlgauss{} outperform \twohot{}, indicating that spreading probability mass to neighbouring locations likely results in less overfitting. Interestingly, we notice that the second hypothesis is also at play, as the optimal value of $\sigma$ seems to be independent of number of bins, indicating that \hlgauss{} generalizes best across a specific range of target values and is indeed leveraging the ordinal nature of the regression problem. Thus, the gains from \hlgauss{} cannot be entirely attributed to overfitting, as is believed to be the case for label smoothing.

\subsection{What Challenges Does Classification Address in Value-Based RL?}

Having seen that the performance gains of cross-entropy losses stem from both the use of a categorical representation of values and distributed targets, we now attempt to understand which challenges in value-based RL cross-entropy losses address, or at least, partially alleviate.

\subsubsection{Is Classification More Robust to Noisy Targets?}
\label{subsec:noisy_targets}

Classification is less prone to overfitting to noisy targets than regression, as it focuses on the categorical relationship between the input and target rather than their exact numerical relationship. We investigate whether classification could better deal with noise induced by stochasticity in RL.

\begin{figure*}[t]
    \centering
    \begin{minipage}[c]{.5\textwidth}
        \centering
        \includegraphics[width=0.9\linewidth]{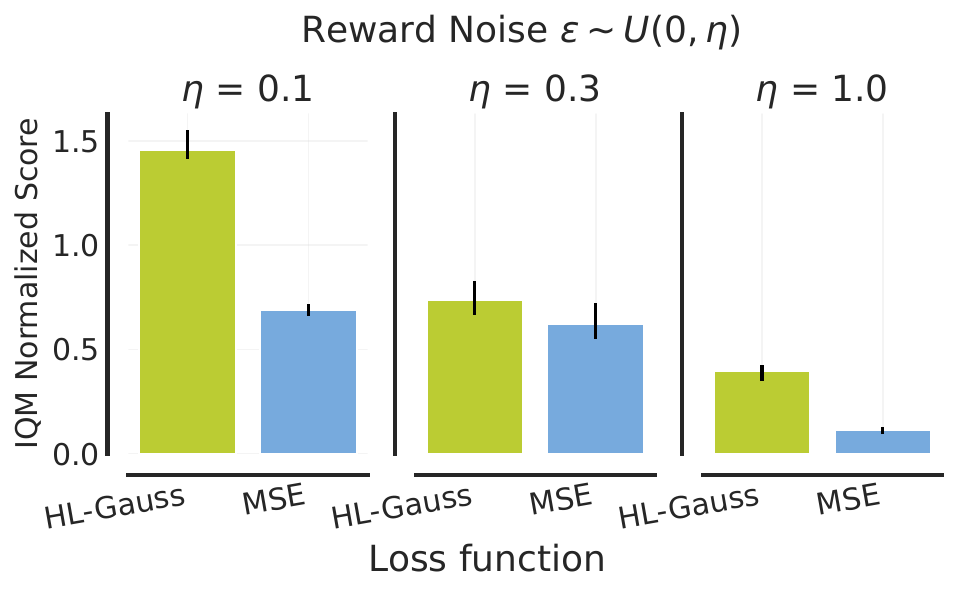}
        \caption{\footnotesize{\textbf{\hlgauss{} vs. MSE when trained using noisy rewards in an offline RL setting on Atari ~(\sref{subsec:single_task_atari}).} Performance of \hlgauss{} degrades slower than MSE as noise increases. Details are in \sref{subsec:noisy_targets}}.}
        \label{fig:noise}
    \end{minipage}\hfill
    \begin{minipage}[c]{0.4\textwidth}
        \centering
        \includegraphics[width=0.9\linewidth]{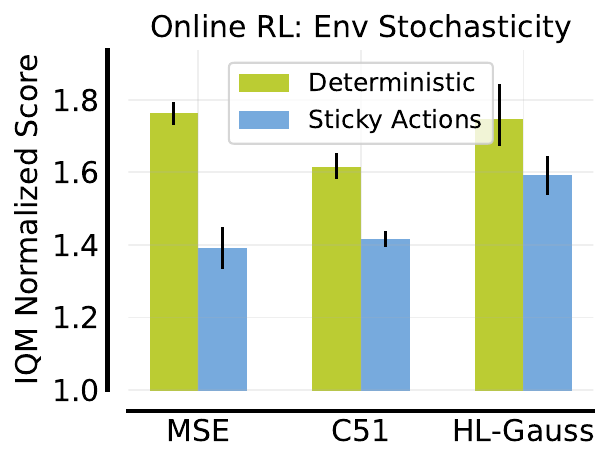}
        \caption{\footnotesize{\textbf{Cross-entropy vs regression losses when varying environment stochasticity in online RL on Atari ~(\sref{subsec:single_task_atari}).} \hlgauss{} only outperforms MSE under deterministic dynamics. Details are in \sref{subsec:noisy_targets}}}.
        \label{fig:atari_sticky_actions}
    \end{minipage}
    \vspace{-0.5cm}
\end{figure*}

\textbf{(a) Noisy Rewards}. To test robustness of classification to stochasticity in rewards, we consider an offline RL setup where we add random noise $\varepsilon_t$, sampled uniformly from $(0, \eta)$, to each dataset reward $r_t$. We vary the noise scale $\eta \in \{0.1, 0.3, 1.0\}$ and compare the performance of cross-entropy based \hlgauss{} with the MSE loss. As shown in Figure~\ref{fig:noise}, the performance of \hlgauss{} degrades more gracefully than MSE as the noise scale increases.

\textbf{(b) Stochasticity in Dynamics}. Following \citet{machado2018revisiting}, our Atari experiments use sticky actions --- with 25\% probability, the environment will execute the previous action again, instead of the agent’s executed action --- resulting in non-deterministic dynamics. 
Here, we turn off sticky actions to compare different losses on deterministic Atari~(60 games). As shown in Figure~\ref{fig:atari_sticky_actions}, while cross-entropy based HL-Gauss outperforms MSE with stochastic dynamics, they perform comparably under deterministic dynamics while outperforming distributional C51.

Overall, the benefits of cross-entropy losses can be partly attributed  to less overfitting to noisy targets, an issue inherent to RL environments with stochastic dynamics or rewards. Such stochasticity issues may also arise as a result of dynamics mis-specification or action delays in real-world embodied RL problems, implying that a cross-entropy loss is a superior choice in those problems.

\subsubsection{Does Classification Learn More Expressive Representations?}
\label{sec:repr_analysis}

It is well known that just using the mean-squared regression error alone does not produce useful representations in value-based RL, often resulting in low capacity representations~\citep{kumar2021implicit} that are incapable of fitting target values observed during subsequent training. Predicting a categorical distribution rather than a scalar target can lead to better representations~\citep{zhang2023improving}, that retain the representational power to model value functions of arbitrary policies that might be encountered over the course of value learning~\citep{dabney21vip}. \citet{lyle19distrl} showed that gains from C51 can be partially attributed to improved representations but it remains unknown whether they stem from backing up distributions of returns or the use of cross-entropy loss.

\begin{figure*}[h]
    \centering
        \includegraphics[width=0.65\linewidth]{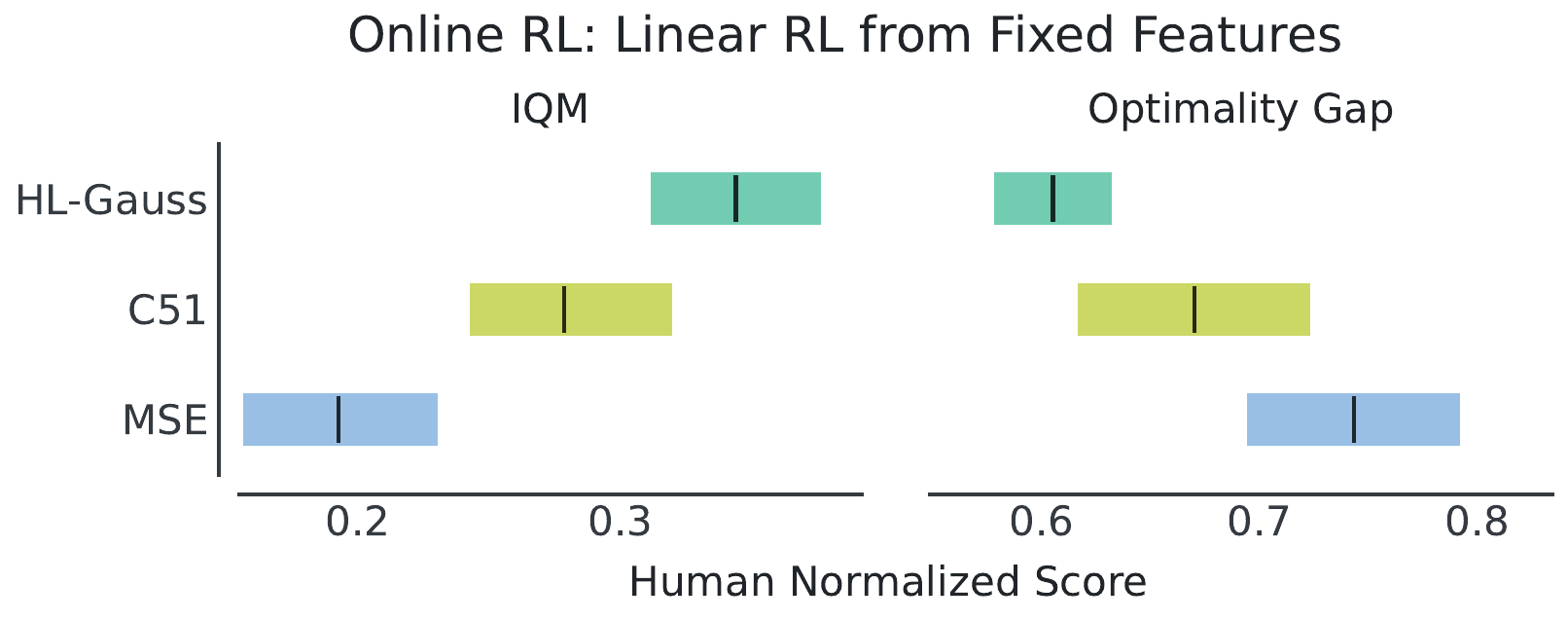}
        \vspace{-0.15cm}
        \caption{\footnotesize{\textbf{Evaluating representations using linear probing ~(\sref{sec:repr_analysis}) on Atari.} This experiment follows the protocol of \citet{farebrother2023pvn}. Optimality gap refers to the distance from human-level performance and lower is better. In both plots, HL-Gauss scores best, indicating its learned representations are the most conducive to downstream tasks.}} 
        \label{fig:linear_rl}
\end{figure*}

To investigate this question, following the protocol in \citet{farebrother2023pvn}, we study whether a learned representation, corresponding to penultimate feature vectors, obtained from value-networks trained online on Atari for 200M frames, still retain the necessary information to re-learn a policy from scratch.
To do so, we train a Q-function with a single linear layer on top of frozen representation~\citep{farebrother2023pvn}, akin to how self-supervised representations are evaluated in  vision~\citep{he2020momentum}.
As shown in Figure~\ref{fig:linear_rl}, cross-entropy losses result in better performance with linear probing. This indicates that their learned representations are indeed better in terms of supporting the value-improvement path of a policy trained from scratch~\citep{dabney21vip}.

\subsubsection{Does Classification Perform Better Amidst Non-Stationarity?}
\label{sec:analysis_nst}

Non-stationarity is inherent to value-based RL as the target computation involves a constantly evolving argmax policy and value function. \citet{bellemare17dist} hypothesized that classification might mitigate difficulty of learning from a non-stationary policy, but did not empirically validate it. Here, we investigate whether classification can indeed handle target non-stationarity better than regression.

\textbf{Synthetic setup}: We first consider a synthetic regression task on CIFAR10 presented in~\citet{lyle2024disentangling}, where the regression target corresponds to mapping an input image $x_i$ through a randomly initialized neural network $f_{\theta^{-}}$ to produce high-frequency targets $y_i = \sin(10^5 \cdot f_{\theta^{-}}(x_i)) + b$ where $b$ is a constant bias that can control for the magnitude of the targets. When learning a value function with TD, the prediction targets are non-stationary and often increase in magnitude over time as the policy improves. We simulate this setting by fitting a network with different losses on the increasing sequence of bias $b \in \{0, 8, 16, 24, 32 \}$. See details in Appendix~\ref{sec:app}. As shown in Figure~\ref{fig:synthetic_magnitude_main}, classification losses retain higher plasticity under non-stationary targets compared to regression. 

\textbf{Offline RL}: To control non-stationarity in an RL context, we run offline SARSA, which estimates the value of the fixed data-collection policy, following the protcol in \citet{kumar2021dr3}. Contrary to Q-learning, which use the action which maximizes the learned Q-value at the next state $S_{t+1}$ for computing the Bellman target~(\sref{sec:rl_back}), SARSA uses the action observed at the next timestep $(S_{t+1}, A_{t+1})$ in the offline dataset. As shown in Figure~\ref{fig:ql_vs_sarsa}, most of the benefit from \hlgauss{} compared to the MSE loss vanishes in the offline SARSA setting, adding evidence that some of the benefits from classification stem from dealing with non-stationarity in value-based RL.

\begin{figure*}[t]
    \centering
    \begin{minipage}[b]{.51\textwidth}
        \centering
        \includegraphics[width=\linewidth]{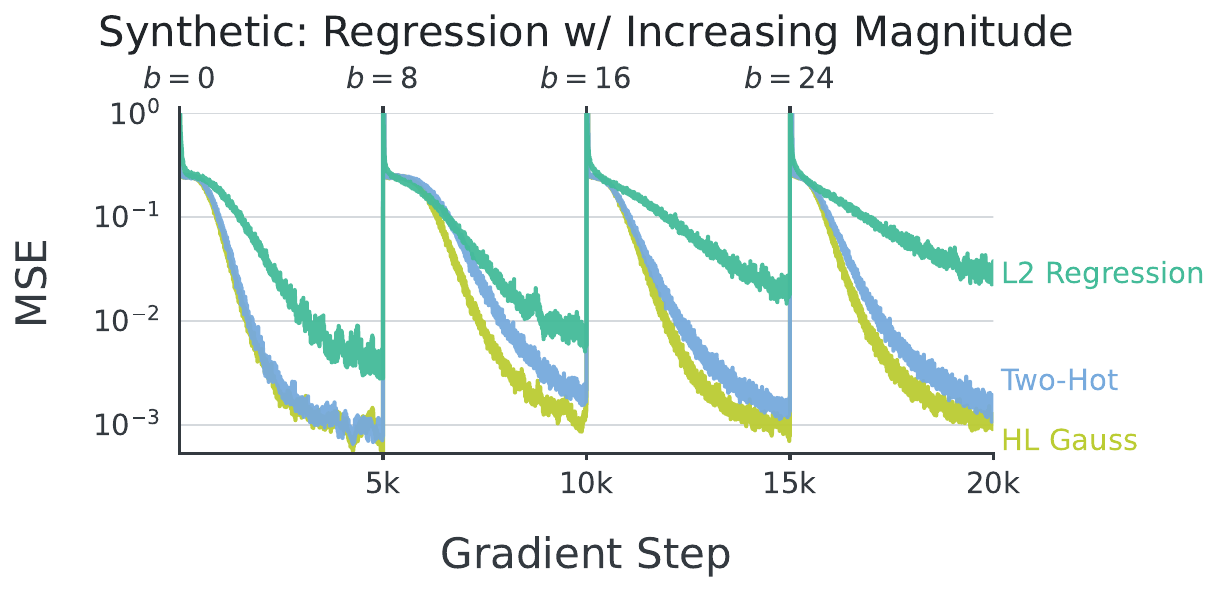}
        \vspace{-0.65cm}
        \caption{\footnotesize{\textbf{Synthetic magnitude prediction experiment to simulate non-stationarity on CIFAR10~(\sref{sec:analysis_nst}).} Non-stationarity is simulated by fitting networks with different losses on an increasing sequences of biases over gradient steps. Cross-entropy losses are less likely to lose plasticity.}}
        \label{fig:synthetic_magnitude_main}
    \end{minipage}
    \hfill
    \begin{minipage}[b]{0.46\textwidth}
        \centering
        \includegraphics[width=\linewidth]{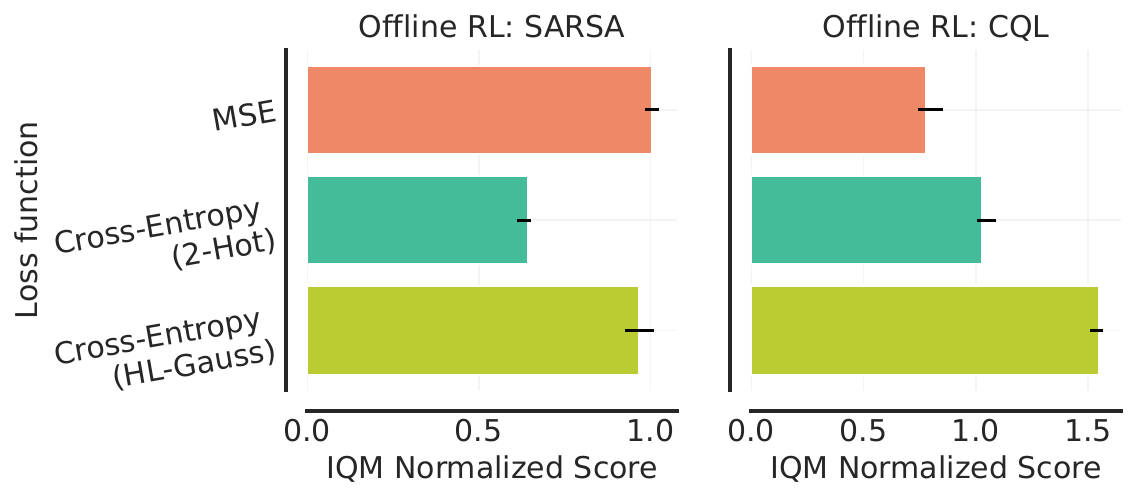}
        \vspace{-0.5cm}
        \caption{\footnotesize{\textbf{Offline QL vs SARSA to ablate policy non-stationarity on Atari~(\sref{sec:analysis_nst}).} HL-Gauss gains over MSE vanish with SARSA. This is evidence that some of the benefits from classification stem from dealing with non-stationarity in value-based RL.}}
        \label{fig:ql_vs_sarsa}
    \end{minipage}
    \vspace{-0.3cm}
\end{figure*}

\textbf{To summarize}, we find that the use of cross-entropy loss itself is central to obtain good performance in value-based RL, and while these methods do not address any specific challenge, they enable value-based RL methods to deal better with non-stationarity, induce highly-expressive representations, and provide robustness against noisy target values.

\section{Related Work}\label{sec:related}

Prior works in tabular regression~\citep{weiss1995rulebased, torgo1996regression} and computer vision~\citep{van2016pixel, kendall2017end, rothe2018deep, rogez2019lcr} have replaced regression with classification to improve performance.
Most notably, \citet{imani2018improving} proposed the \hlgauss{} cross-entropy loss for regression and show its efficacy on small-scale supervised regression tasks, outside of RL. 
Our work complements these prior works by illustrating for the first time that a classification objective trained with cross-entropy, particularly \hlgauss{}, can enable effectively scaling for value-based RL on a variety of domains, including Atari, robotic manipulation, chess, and Wordle.

Several state-of-the-art methods in RL have used the \twohot{} cross-entropy loss without any analysis, either as an ``ad-hoc'' trick ~\citep{schrittwieser20muzero}, citing benefits for sparse rewards~\citep{hafner23dreamerv3}, or simply relying on folk wisdom~\citep{hessel21museli, hansen23tdmpc2}.
However, in our experiments, \twohot{} performs worse than other cross-entropy losses and MSE. We believe this is because \twohot{} does not effectively distribute probability to neighboring classes, unlike C51 and \hlgauss~(see \sref{sec:distributed_targets} for an empirical investigation).

Closely related is the line of work on categorical distributional RL. 
Notably, \citet{achab23onestep} offer an analysis of categorical one-step distributional RL, which corresponds precisely to the \twohot{} algorithm discussed herein with the similarity of these two approaches not being previously recognized.
Additionally, the work of \citet{bellemare17dist} pioneered the C51 algorithm, and while their primary focus \emph{was not} on framing RL as classification, our findings suggest that the specific loss function employed may play a more significant role in the algorithm's success than modeling the return distribution itself.
Several methods find that categorical distributional RL losses are important for scaling offline value-based RL \citep{kumar2022offline, springenberg2024offline}, but these works do not attempt to isolate which components of this paradigm are crucial for attaining positive scaling trends.
We also note that these findings do not contradict recent theoretical work~\citep{wang2023drlbounds, rowland23qtd} which argues that distributional RL brings statistical benefits over standard RL orthogonal to use of a cross entropy objective or the categorical representation.

Prior works have characterized the representations learned by TD-learning~\citep{bellemare2019avf, lyle21auxtask, lelan22generalization, lelan23bootstrap,kumar2021implicit,kumar2021dr3} but these prior works focus entirely on MSE losses with little to no work analyzing representations learned by cross-entropy based losses in RL. Our linear probing experiments in \sref{sec:repr_analysis} try to fill this void, demonstrating that value-networks trained with cross-entropy losses learn better representations than regression. This finding is especially important since \citet{imani2018improving} did not find any representational benefits of \hlgauss{} over MSE on supervised regression, indicating that the use of cross-entropy might have substantial benefits for TD-based learning methods in particular.

\section{Conclusion}

In this paper, we showed that framing regression as classification and minimizing categorical cross-entropy instead of the mean squared error yields large improvements in performance and  scalability of value-based RL methods, on a wide variety of tasks, with several neural network architectures. We analyzed the source of these improvements and found that they stem specifically from the ability of the cross-entropy loss in enabling more expressive representations and handling noise and non-stationarity in value-based RL better. While the cross-entropy loss alone does not fully alleviate any of these problems entirely, our results show the substantial difference this small change can make.

We believe that strong results with the use categorical cross-entropy has implications for future algorithm design in deep RL, both in theory and practice. For instance, value-based RL approaches have been harder to scale and tune when the value function is represented by a transformer architecture and our results hint that classification might provide for a smooth approach to translate innovation in value-based RL to transformers.
From a theoretical perspective, analyzing the optimization dynamics of cross-entropy might help devise improved losses or target distribution representations.
Finally, while we did explore a number of settings, further work is required to evaluate the efficacy of classification losses in other RL problems such as those involving pre-training, fine-tuning, or continual RL.

\section*{Acknowledgements}

We would like to thank Will Dabney for providing feedback on an early version of this paper. We'd also like to thank Clare Lyle, Mark Rowland, Marc Bellemare, Max Schwarzer, Pierluca D'oro, Nate Rahn, Harley Wiltzer, Wesley Chung, and Dale Schuurmans, for informative discussions. We'd also like to acknowledge Anian Ruoss, Grégoire Delétang, and Tim Genewein for their help with the Chess training infrastructure. This research was supported by the TPU resources at Google DeepMind, and the authors are grateful to Doina Precup and Joelle Baral for their support.

\section*{Author Contributions}

JF led the project, implemented histogram-based methods, ran all the single-task online RL experiments on Atari, Q-distillation on Chess, jointly proposed and ran most of the analysis experiments, and contributed significantly to paper writing.

JO and AAT set up and ran the multi-task RL experiments and helped with writing. QV ran the robotic manipulation experiments and YC helped with the initial set-up. TX helped with paper writing and AI was involved in discussions. SL advised on the robotics and Wordle experiments and provided feedback. PSC helped set up the SoftMoE experiments and hosted Jesse at GDM. PSC and AF sponsored the project and took part in discussions.

AK advised the project, proposed offline RL analysis for non-stationarity and representation learning, contributed significantly to writing, revising, and the narrative, and set up the robotics and multi-game scaling experiments.
RA proposed the research direction, advised the project, led the paper writing, ran offline RL and Wordle experiments, and helped set up all of the multi-task scaling and non-Atari experiments.

{
\hypersetup{urlcolor=black}%
\bibliography{references}%

\begin{thebibliography}{63}
\providecommand{\natexlab}[1]{#1}
\providecommand{\url}[1]{\texttt{#1}}
\expandafter\ifx\csname urlstyle\endcsname\relax
  \providecommand{\doi}[1]{doi: #1}\else
  \providecommand{\doi}{doi: \begingroup \urlstyle{rm}\Url}\fi

\bibitem[Achab et~al.(2023)Achab, Alami, Djilali, Fedyanin, and
  Moulines]{achab23onestep}
Mastane Achab, R{\'{e}}da Alami, Yasser Abdelaziz~Dahou Djilali, Kirill
  Fedyanin, and Eric Moulines.
\newblock One-step distributional reinforcement learning.
\newblock \emph{CoRR}, abs/2304.14421, 2023.

\bibitem[Agarwal et~al.(2020)Agarwal, Schuurmans, and
  Norouzi]{agarwal2020optimistic}
Rishabh Agarwal, Dale Schuurmans, and Mohammad Norouzi.
\newblock An optimistic perspective on offline reinforcement learning.
\newblock In \emph{International Conference on Machine Learning (ICML)}, 2020.

\bibitem[Agarwal et~al.(2021)Agarwal, Schwarzer, Castro, Courville, and
  Bellemare]{agarwal2021precipice}
Rishabh Agarwal, Max Schwarzer, Pablo~Samuel Castro, Aaron Courville, and
  Marc~G. Bellemare.
\newblock Deep reinforcement learning at the edge of the statistical precipice.
\newblock \emph{Neural Information Processing Systems (NeurIPS)}, 2021.

\bibitem[Ali~Ta\"{i}ga et~al.(2023)Ali~Ta\"{i}ga, Agarwal, Farebrother,
  Courville, and Bellemare]{taiga2023multitask}
Adrien Ali~Ta\"{i}ga, Rishabh Agarwal, Jesse Farebrother, Aaron Courville, and
  Marc~G. Bellemare.
\newblock Investigating multi-task pretraining and generalization in
  reinforcement learning.
\newblock In \emph{International Conference on Learning Representations
  (ICLR)}, 2023.

\bibitem[Bellemare et~al.(2013)Bellemare, Naddaf, Veness, and
  Bowling]{bellemare2013arcade}
Marc~G. Bellemare, Yavar Naddaf, Joel Veness, and Michael Bowling.
\newblock The arcade learning environment: An evaluation platform for general
  agents.
\newblock \emph{Journal of Artificial Intelligence Research (JAIR)},
  47:\penalty0 253--279, 2013.

\bibitem[Bellemare et~al.(2017)Bellemare, Dabney, and Munos]{bellemare17dist}
Marc~G. Bellemare, Will Dabney, and R{\'{e}}mi Munos.
\newblock A distributional perspective on reinforcement learning.
\newblock In \emph{International Conference on Machine Learning (ICML)}, 2017.

\bibitem[Bellemare et~al.(2019)Bellemare, Dabney, Dadashi, Ali~Ta\"{i}ga,
  Castro, Le~Roux, Schuurmans, Lattimore, and Lyle]{bellemare2019avf}
Marc~G. Bellemare, Will Dabney, Robert Dadashi, Adrien Ali~Ta\"{i}ga,
  Pablo~Samuel Castro, Nicolas Le~Roux, Dale Schuurmans, Tor Lattimore, and
  Clare Lyle.
\newblock A geometric perspective on optimal representations for reinforcement
  learning.
\newblock In \emph{Neural Information Processing Systems (NeurIPS)}, 2019.

\bibitem[Bellemare et~al.(2023)Bellemare, Dabney, and
  Rowland]{bellemare23distbook}
Marc~G. Bellemare, Will Dabney, and Mark Rowland.
\newblock \emph{Distributional reinforcement learning}.
\newblock MIT Press, 2023.

\bibitem[Bradbury et~al.(2018)Bradbury, Frostig, Hawkins, Johnson, Leary,
  Maclaurin, Necula, Paszke, Vander{P}las, Wanderman-{M}ilne, and
  Zhang]{jax2018github}
James Bradbury, Roy Frostig, Peter Hawkins, Matthew~James Johnson, Chris Leary,
  Dougal Maclaurin, George Necula, Adam Paszke, Jake Vander{P}las, Skye
  Wanderman-{M}ilne, and Qiao Zhang.
\newblock {JAX}: composable transformations of {P}ython+{N}um{P}y programs,
  2018.
\newblock URL \url{http://github.com/google/jax}.

\bibitem[Castro et~al.(2018)Castro, Moitra, Gelada, Kumar, and
  Bellemare]{castro18dopamine}
Pablo~Samuel Castro, Subhodeep Moitra, Carles Gelada, Saurabh Kumar, and
  Marc~G. Bellemare.
\newblock Dopamine: {A} {R}esearch {F}ramework for {D}eep {R}einforcement
  {L}earning.
\newblock \emph{CoRR}, abs/1812.06110, 2018.

\bibitem[Ceron and Castro(2021)]{ceron2021revisiting}
Johan Samir~Obando Ceron and Pablo~Samuel Castro.
\newblock Revisiting rainbow: Promoting more insightful and inclusive deep
  reinforcement learning research.
\newblock In \emph{International Conference on Machine Learning (ICML)}, 2021.

\bibitem[Chebotar et~al.(2023)Chebotar, Vuong, Hausman, Xia, Lu, Irpan, Kumar,
  Yu, Herzog, Pertsch, et~al.]{chebotar2023q}
Yevgen Chebotar, Quan Vuong, Karol Hausman, Fei Xia, Yao Lu, Alex Irpan, Aviral
  Kumar, Tianhe Yu, Alexander Herzog, Karl Pertsch, et~al.
\newblock Q-transformer: Scalable offline reinforcement learning via
  autoregressive q-functions.
\newblock In \emph{Conference on Robot Learning (CoRL)}, 2023.

\bibitem[Dabney et~al.(2021)Dabney, Barreto, Rowland, Dadashi, Quan, Bellemare,
  and Silver]{dabney21vip}
Will Dabney, André Barreto, Mark Rowland, Robert Dadashi, John Quan, Marc~G.
  Bellemare, and David Silver.
\newblock The value-improvement path: Towards better representations for
  reinforcement learning.
\newblock In \emph{AAAI Conference on Artificial Intelligence}, 2021.

\bibitem[Espeholt et~al.(2018)Espeholt, Soyer, Munos, Simonyan, Mnih, Ward,
  Doron, Firoiu, Harley, Dunning, Legg, and Kavukcuoglu]{espeholt2018impala}
Lasse Espeholt, Hubert Soyer, Remi Munos, Karen Simonyan, Vlad Mnih, Tom Ward,
  Yotam Doron, Vlad Firoiu, Tim Harley, Iain Dunning, Shane Legg, and Koray
  Kavukcuoglu.
\newblock Impala: Scalable distributed deep-rl with importance weighted
  actor-learner architectures.
\newblock In \emph{International Conference on Machine Learning (ICML)}, 2018.

\bibitem[Farebrother et~al.(2018)Farebrother, Machado, and
  Bowling]{farebrother18generalization}
Jesse Farebrother, Marlos~C. Machado, and Michael Bowling.
\newblock Generalization and regularization in {DQN}.
\newblock \emph{CoRR}, abs/1810.00123, 2018.

\bibitem[Farebrother et~al.(2023)Farebrother, Greaves, Agarwal, Le~Lan,
  Goroshin, Samuel~Castro, and Bellemare]{farebrother2023pvn}
Jesse Farebrother, Joshua Greaves, Rishabh Agarwal, Charline Le~Lan, Ross
  Goroshin, Pablo Samuel~Castro, and Marc~G. Bellemare.
\newblock Proto-value networks: Scaling representation learning with auxiliary
  tasks.
\newblock In \emph{International Conference on Learning Representations
  (ICLR)}, 2023.

\bibitem[Hafner et~al.(2023)Hafner, Pasukonis, Ba, and
  Lillicrap]{hafner23dreamerv3}
Danijar Hafner, Jurgis Pasukonis, Jimmy Ba, and Timothy~P. Lillicrap.
\newblock Mastering diverse domains through world models.
\newblock \emph{CoRR}, abs/2301.04104, 2023.

\bibitem[Hansen et~al.(2024)Hansen, Su, and Wang]{hansen23tdmpc2}
Nicklas Hansen, Hao Su, and Xiaolong Wang.
\newblock {TD}-{MPC}2: Scalable, robust world models for continuous control.
\newblock In \emph{International Conference on Learning Representations
  (ICLR)}, 2024.

\bibitem[He et~al.(2016)He, Zhang, Ren, and Sun]{resnet}
Kaiming He, Xiangyu Zhang, Shaoqing Ren, and Jian Sun.
\newblock Deep residual learning for image recognition.
\newblock In \emph{IEEE Conference on Computer Vision and Pattern Recognition
  (CVPR)}, 2016.

\bibitem[He et~al.(2020)He, Fan, Wu, Xie, and Girshick]{he2020momentum}
Kaiming He, Haoqi Fan, Yuxin Wu, Saining Xie, and Ross Girshick.
\newblock Momentum contrast for unsupervised visual representation learning.
\newblock In \emph{IEEE Conference on Computer Vision and Pattern Recognition
  (CVPR)}, 2020.

\bibitem[Hessel et~al.(2021)Hessel, Danihelka, Viola, Guez, Schmitt, Sifre,
  Weber, Silver, and van Hasselt]{hessel21museli}
Matteo Hessel, Ivo Danihelka, Fabio Viola, Arthur Guez, Simon Schmitt, Laurent
  Sifre, Theophane Weber, David Silver, and Hado van Hasselt.
\newblock Muesli: Combining improvements in policy optimization.
\newblock In \emph{International Conference on Machine Learning {(ICML)}},
  2021.

\bibitem[Ho et~al.(2021)Ho, Rao, Xu, Jang, Khansari, and Bai]{ho2020retinagan}
Daniel Ho, Kanishka Rao, Zhuo Xu, Eric Jang, Mohi Khansari, and Yunfei Bai.
\newblock Retinagan: An object-aware approach to sim-to-real transfer.
\newblock In \emph{IEEE International Conference on Robotics and Automation
  (ICRA)}, 2021.

\bibitem[Imani and White(2018)]{imani2018improving}
Ehsan Imani and Martha White.
\newblock Improving regression performance with distributional losses.
\newblock In \emph{International Conference on Machine Learning (ICML)}, 2018.

\bibitem[Imani et~al.(2024)Imani, Luedemann, Scholnick-Hughes, Elelimy, and
  White]{imani2024investigating}
Ehsan Imani, Kai Luedemann, Sam Scholnick-Hughes, Esraa Elelimy, and Martha
  White.
\newblock Investigating the histogram loss in regression.
\newblock \emph{CoRR}, abs/2402.13425, 2024.

\bibitem[Kaplan et~al.(2020)Kaplan, McCandlish, Henighan, Brown, Chess, Child,
  Gray, Radford, Wu, and Amodei]{kaplan2020scaling}
Jared Kaplan, Sam McCandlish, Tom Henighan, Tom~B Brown, Benjamin Chess, Rewon
  Child, Scott Gray, Alec Radford, Jeffrey Wu, and Dario Amodei.
\newblock Scaling laws for neural language models.
\newblock \emph{CoRR}, abs/2001.08361, 2020.

\bibitem[Kendall et~al.(2017)Kendall, Martirosyan, Dasgupta, Henry, Kennedy,
  Bachrach, and Bry]{kendall2017end}
Alex Kendall, Hayk Martirosyan, Saumitro Dasgupta, Peter Henry, Ryan Kennedy,
  Abraham Bachrach, and Adam Bry.
\newblock End-to-end learning of geometry and context for deep stereo
  regression.
\newblock In \emph{IEEE International Conference on Computer Vision (ICCV)},
  2017.

\bibitem[Krizhevsky et~al.(2012)Krizhevsky, Sutskever, and
  Hinton]{krizhevsky2012imagenet}
Alex Krizhevsky, Ilya Sutskever, and Geoffrey~E Hinton.
\newblock Imagenet classification with deep convolutional neural networks.
\newblock \emph{Neural Information Processing Systems (NeurIPS)}, 2012.

\bibitem[Kumar et~al.(2020)Kumar, Zhou, Tucker, and
  Levine]{kumar2020conservative}
Aviral Kumar, Aurick Zhou, George Tucker, and Sergey Levine.
\newblock Conservative q-learning for offline reinforcement learning.
\newblock \emph{Neural Information Processing Systems (NeurIPS)}, 2020.

\bibitem[Kumar et~al.(2021)Kumar, Agarwal, Ghosh, and
  Levine]{kumar2021implicit}
Aviral Kumar, Rishabh Agarwal, Dibya Ghosh, and Sergey Levine.
\newblock Implicit under-parameterization inhibits data-efficient deep
  reinforcement learning.
\newblock In \emph{International Conference on Learning Representations
  (ICLR)}, 2021.

\bibitem[Kumar et~al.(2022)Kumar, Agarwal, Ma, Courville, Tucker, and
  Levine]{kumar2021dr3}
Aviral Kumar, Rishabh Agarwal, Tengyu Ma, Aaron Courville, George Tucker, and
  Sergey Levine.
\newblock Dr3: Value-based deep reinforcement learning requires explicit
  regularization.
\newblock In \emph{International Conference on Learning Representations
  (ICLR)}, 2022.

\bibitem[Kumar et~al.(2023)Kumar, Agarwal, Geng, Tucker, and
  Levine]{kumar2022offline}
Aviral Kumar, Rishabh Agarwal, Xinyang Geng, George Tucker, and Sergey Levine.
\newblock {Offline Q-Learning on Diverse Multi-Task Data Both Scales and
  Generalizes}.
\newblock In \emph{International Conference on Learning Representations
  (ICLR)}, 2023.

\bibitem[Le~Lan et~al.(2022)Le~Lan, Tu, Oberman, Agarwal, and
  Bellemare]{lelan22generalization}
Charline Le~Lan, Stephen Tu, Adam Oberman, Rishabh Agarwal, and Marc~G.
  Bellemare.
\newblock On the generalization of representations in reinforcement learning.
\newblock In \emph{International Conference on Artificial Intelligence and
  Statistics (AISTATS)}, 2022.

\bibitem[Le~Lan et~al.(2023)Le~Lan, Tu, Rowland, Harutyunyan, Agarwal,
  Bellemare, and Dabney]{lelan23bootstrap}
Charline Le~Lan, Stephen Tu, Mark Rowland, Anna Harutyunyan, Rishabh Agarwal,
  Marc~G. Bellemare, and Will Dabney.
\newblock Bootstrapped representations in reinforcement learning.
\newblock In \emph{International Conference on Machine Learning (ICML)}, 2023.

\bibitem[Lee et~al.(2022)Lee, Nachum, Yang, Lee, Freeman, Guadarrama, Fischer,
  Xu, Jang, Michalewski, and Mordatch]{lee2022multigame}
Kuang-Huei Lee, Ofir Nachum, Mengjiao~(Sherry) Yang, Lisa Lee, Daniel Freeman,
  Sergio Guadarrama, Ian Fischer, Winnie Xu, Eric Jang, Henryk Michalewski, and
  Igor Mordatch.
\newblock Multi-game decision transformers.
\newblock In \emph{Neural Information Processing Systems (NeurIPS)}, 2022.

\bibitem[Lehnert et~al.(2024)Lehnert, Sukhbaatar, Mcvay, Rabbat, and
  Tian]{lehnert2024beyond}
Lucas Lehnert, Sainbayar Sukhbaatar, Paul Mcvay, Michael Rabbat, and Yuandong
  Tian.
\newblock Beyond a*: Better planning with transformers via search dynamics
  bootstrapping.
\newblock \emph{CoRR}, abs/2402.14083, 2024.

\bibitem[Levine et~al.(2020)Levine, Kumar, Tucker, and Fu]{levine2020offline}
Sergey Levine, Aviral Kumar, George Tucker, and Justin Fu.
\newblock {Offline Reinforcement Learning: Tutorial, Review, and Perspectives
  on Open Problems}.
\newblock \emph{CoRR}, abs/2005.01643, 2020.

\bibitem[Lyle et~al.(2019)Lyle, Bellemare, and Castro]{lyle19distrl}
Clare Lyle, Marc~G. Bellemare, and Pablo~Samuel Castro.
\newblock A comparative analysis of expected and distributional reinforcement
  learning.
\newblock In \emph{AAAI Conference on Artificial Intelligence}, 2019.

\bibitem[Lyle et~al.(2021)Lyle, Rowland, Ostrovski, and Dabney]{lyle21auxtask}
Clare Lyle, Mark Rowland, Georg Ostrovski, and Will Dabney.
\newblock On the effect of auxiliary tasks on representation dynamics.
\newblock In \emph{International Conference on Artificial Intelligence and
  Statistics (AISTATS)}, 2021.

\bibitem[Lyle et~al.(2022)Lyle, Rowland, and Dabney]{lyle2022understanding}
Clare Lyle, Mark Rowland, and Will Dabney.
\newblock Understanding and preventing capacity loss in reinforcement learning.
\newblock In \emph{International Conference on Learning Representations
  (ICLR)}, 2022.

\bibitem[Lyle et~al.(2024)Lyle, Zheng, Khetarpal, van Hasselt, Pascanu,
  Martens, and Dabney]{lyle2024disentangling}
Clare Lyle, Zeyu Zheng, Khimya Khetarpal, Hado van Hasselt, Razvan Pascanu,
  James Martens, and Will Dabney.
\newblock Disentangling the causes of plasticity loss in neural networks.
\newblock \emph{CoRR}, abs/2402.18762, 2024.

\bibitem[Machado et~al.(2018)Machado, Bellemare, Talvitie, Veness, Hausknecht,
  and Bowling]{machado2018revisiting}
Marlos~C. Machado, Marc~G. Bellemare, Erik Talvitie, Joel Veness, Matthew
  Hausknecht, and Michael Bowling.
\newblock Revisiting the arcade learning environment: Evaluation protocols and
  open problems for general agents.
\newblock \emph{Journal of Artificial Intelligence Research (JAIR)},
  61:\penalty0 523--562, 2018.

\bibitem[Mnih et~al.(2015)Mnih, Kavukcuoglu, Silver, Rusu, Veness, Bellemare,
  Graves, Riedmiller, Fidjeland, Ostrovski, Petersen, Beattie, Sadik,
  Antonoglou, King, Kumaran, Wierstra, Legg, and Hassabis]{mnih15dqn}
Volodymyr Mnih, Koray Kavukcuoglu, David Silver, Andrei~A. Rusu, Joel Veness,
  Marc~G. Bellemare, Alex Graves, Martin Riedmiller, Andreas~K. Fidjeland,
  Georg Ostrovski, Stig Petersen, Charles Beattie, Amir Sadik, Ioannis
  Antonoglou, Helen King, Dharshan Kumaran, Daan Wierstra, Shane Legg, and
  Demis Hassabis.
\newblock Human-level control through deep reinforcement learning.
\newblock \emph{Nature}, 518\penalty0 (7540):\penalty0 529–533, 2015.

\bibitem[Mnih et~al.(2016)Mnih, Badia, Mirza, Graves, Lillicrap, Harley,
  Silver, and Kavukcuoglu]{mnih2016asynchronous}
Volodymyr Mnih, Adria~Puigdomenech Badia, Mehdi Mirza, Alex Graves, Timothy
  Lillicrap, Tim Harley, David Silver, and Koray Kavukcuoglu.
\newblock Asynchronous methods for deep reinforcement learning.
\newblock In \emph{International Conference on Machine Learning (ICML)}, 2016.

\bibitem[Obando-Ceron et~al.(2024)Obando-Ceron, Sokar, Willi, Lyle,
  Farebrother, Foerster, Dziugaite, Precup, and Castro]{obando2024mixtures}
Johan Obando-Ceron, Ghada Sokar, Timon Willi, Clare Lyle, Jesse Farebrother,
  Jakob Foerster, Gintare~Karolina Dziugaite, Doina Precup, and Pablo~Samuel
  Castro.
\newblock Mixtures of experts unlock parameter scaling for deep rl.
\newblock \emph{CoRR}, abs/2402.08609, 2024.

\bibitem[Paszke et~al.(2019)Paszke, Gross, Massa, Lerer, Bradbury, Chanan,
  Killeen, Lin, Gimelshein, Antiga, Desmaison, Kopf, Yang, DeVito, Raison,
  Tejani, Chilamkurthy, Steiner, Fang, Bai, and Chintala]{pytorch}
Adam Paszke, Sam Gross, Francisco Massa, Adam Lerer, James Bradbury, Gregory
  Chanan, Trevor Killeen, Zeming Lin, Natalia Gimelshein, Luca Antiga, Alban
  Desmaison, Andreas Kopf, Edward Yang, Zachary DeVito, Martin Raison, Alykhan
  Tejani, Sasank Chilamkurthy, Benoit Steiner, Lu~Fang, Junjie Bai, and Soumith
  Chintala.
\newblock Pytorch: An imperative style, high-performance deep learning library.
\newblock In \emph{Neural Information Processing Systems (NeurIPS)}, 2019.

\bibitem[Pintea et~al.(2023)Pintea, Lin, Dijkstra, and van
  Gemert]{pintea2023step}
Silvia~L. Pintea, Yancong Lin, Jouke Dijkstra, and Jan~C. van Gemert.
\newblock A step towards understanding why classification helps regression.
\newblock In \emph{IEEE International Conference on Computer Vision (ICCV)},
  pages 19972--19981, 2023.

\bibitem[Puigcerver et~al.(2024)Puigcerver, Ruiz, Mustafa, and
  Houlsby]{puigcerver2023sparse}
Joan Puigcerver, Carlos~Riquelme Ruiz, Basil Mustafa, and Neil Houlsby.
\newblock From sparse to soft mixtures of experts.
\newblock In \emph{International Conference on Learning Representations
  (ICLR)}, 2024.

\bibitem[Rogez et~al.(2019)Rogez, Weinzaepfel, and Schmid]{rogez2019lcr}
Gregory Rogez, Philippe Weinzaepfel, and Cordelia Schmid.
\newblock Lcr-net++: Multi-person 2d and 3d pose detection in natural images.
\newblock \emph{IEEE Transactions on Pattern Analysis and Machine Intelligence
  (PAMI)}, 42\penalty0 (5):\penalty0 1146--1161, 2019.

\bibitem[Rothe et~al.(2018)Rothe, Timofte, and Van~Gool]{rothe2018deep}
Rasmus Rothe, Radu Timofte, and Luc Van~Gool.
\newblock Deep expectation of real and apparent age from a single image without
  facial landmarks.
\newblock \emph{International Journal of Computer Vision (IJCV)}, 126\penalty0
  (2-4):\penalty0 144--157, 2018.

\bibitem[Rowland et~al.(2023)Rowland, Tang, Lyle, Munos, Bellemare, and
  Dabney]{rowland23qtd}
Mark Rowland, Yunhao Tang, Clare Lyle, R{\'{e}}mi Munos, Marc~G. Bellemare, and
  Will Dabney.
\newblock The statistical benefits of quantile temporal-difference learning for
  value estimation.
\newblock In \emph{International Conference on Machine Learning ({ICML})},
  2023.

\bibitem[Ruoss et~al.(2024)Ruoss, Del{\'e}tang, Medapati, Grau-Moya, Wenliang,
  Catt, Reid, and Genewein]{ruoss2024grandmaster}
Anian Ruoss, Gr{\'e}goire Del{\'e}tang, Sourabh Medapati, Jordi Grau-Moya,
  Li~Kevin Wenliang, Elliot Catt, John Reid, and Tim Genewein.
\newblock Grandmaster-level chess without search.
\newblock \emph{CoRR}, abs/2402.04494, 2024.

\bibitem[Schrittwieser et~al.(2020)Schrittwieser, Antonoglou, Hubert, Simonyan,
  Sifre, Schmitt, Guez, Lockhart, Hassabis, Graepel, Lillicrap, and
  Silver]{schrittwieser20muzero}
Julian Schrittwieser, Ioannis Antonoglou, Thomas Hubert, Karen Simonyan,
  Laurent Sifre, Simon Schmitt, Arthur Guez, Edward Lockhart, Demis Hassabis,
  Thore Graepel, Timothy~P. Lillicrap, and David Silver.
\newblock Mastering atari, go, chess and shogi by planning with a learned
  model.
\newblock \emph{Nature}, 588\penalty0 (7839):\penalty0 604--609, 2020.

\bibitem[Silver et~al.(2017)Silver, Schrittwieser, Simonyan, Antonoglou, Huang,
  Guez, Hubert, Baker, Lai, Bolton, Chen, Lillicrap, Hui, Sifre, van~den
  Driessche, Graepel, and Hassabis]{silver2017mastering}
David Silver, Julian Schrittwieser, Karen Simonyan, Ioannis Antonoglou, Aja
  Huang, Arthur Guez, Thomas Hubert, Lucas Baker, Matthew Lai, Adrian Bolton,
  Yutian Chen, Timothy Lillicrap, Fan Hui, Laurent Sifre, George van~den
  Driessche, Thore Graepel, and Demis Hassabis.
\newblock Mastering the game of go without human knowledge.
\newblock \emph{Nature}, 550\penalty0 (7676):\penalty0 354–359, 2017.

\bibitem[Snell et~al.(2023)Snell, Kostrikov, Su, Yang, and
  Levine]{snell2022offline}
Charlie~Victor Snell, Ilya Kostrikov, Yi~Su, Sherry Yang, and Sergey Levine.
\newblock Offline {RL} for natural language generation with implicit language q
  learning.
\newblock In \emph{International Conference on Learning Representations
  (ICLR)}, 2023.

\bibitem[Springenberg et~al.(2024)Springenberg, Abdolmaleki, Zhang, Groth,
  Bloesch, Lampe, Brakel, Bechtle, Kapturowski, Hafner,
  et~al.]{springenberg2024offline}
Jost~Tobias Springenberg, Abbas Abdolmaleki, Jingwei Zhang, Oliver Groth,
  Michael Bloesch, Thomas Lampe, Philemon Brakel, Sarah Bechtle, Steven
  Kapturowski, Roland Hafner, et~al.
\newblock Offline actor-critic reinforcement learning scales to large models.
\newblock \emph{CoRR}, abs/2402.05546, 2024.

\bibitem[Stewart et~al.(2023)Stewart, Bach, Berthet, and
  Vert]{stewart2023regression}
Lawrence Stewart, Francis Bach, Quentin Berthet, and Jean-Philippe Vert.
\newblock Regression as classification: Influence of task formulation on neural
  network features.
\newblock In \emph{International Conference on Artificial Intelligence and
  Statistics (AISTATS)}, 2023.

\bibitem[Szegedy et~al.(2016)Szegedy, Vanhoucke, Ioffe, Shlens, and
  Wojna]{szegedy16inception}
Christian Szegedy, Vincent Vanhoucke, Sergey Ioffe, Jonathon Shlens, and
  Zbigniew Wojna.
\newblock Rethinking the inception architecture for computer vision.
\newblock In \emph{{IEEE} Conference on Computer Vision and Pattern Recognition
  ({CVPR})}, 2016.

\bibitem[Torgo and Gama(1996)]{torgo1996regression}
Lu{\'\i}s Torgo and Jo{\~a}o Gama.
\newblock Regression by classification.
\newblock In \emph{Brazilian Symposium on Artificial Intelligence}, pages
  51--60. Springer, 1996.

\bibitem[Van Den~Oord et~al.(2016)Van Den~Oord, Kalchbrenner, and
  Kavukcuoglu]{van2016pixel}
A{\"a}ron Van Den~Oord, Nal Kalchbrenner, and Koray Kavukcuoglu.
\newblock Pixel recurrent neural networks.
\newblock In \emph{International Conference on Machine Learning (ICML)}, 2016.

\bibitem[Vaswani et~al.(2017)Vaswani, Shazeer, Parmar, Uszkoreit, Jones, Gomez,
  Kaiser, and Polosukhin]{vaswani2017attention}
Ashish Vaswani, Noam Shazeer, Niki Parmar, Jakob Uszkoreit, Llion Jones,
  Aidan~N Gomez, {\L}ukasz Kaiser, and Illia Polosukhin.
\newblock Attention is all you need.
\newblock \emph{Neural Information Processing Systems (NeurIPS)}, 2017.

\bibitem[Wang et~al.(2023)Wang, Zhou, Wu, Kallus, and Sun]{wang2023drlbounds}
Kaiwen Wang, Kevin Zhou, Runzhe Wu, Nathan Kallus, and Wen Sun.
\newblock The benefits of being distributional: Small-loss bounds for
  reinforcement learning.
\newblock In \emph{Neural Information Processing Systems (NeurIPS)}, 2023.

\bibitem[Weiss and Indurkhya(1995)]{weiss1995rulebased}
Sholom~M Weiss and Nitin Indurkhya.
\newblock Rule-based machine learning methods for functional prediction.
\newblock \emph{Journal of Artificial Intelligence Research (JAIR)},
  3:\penalty0 383--403, 1995.

\bibitem[Zhang et~al.(2023)Zhang, Yang, Mi, Zheng, and Yao]{zhang2023improving}
Shihao Zhang, Linlin Yang, Michael~Bi Mi, Xiaoxu Zheng, and Angela Yao.
\newblock Improving deep regression with ordinal entropy.
\newblock In \emph{International Conference on Learning Representations
  (ICLR)}, 2023.

\end{thebibliography}
}

\newpage
\appendix
\onecolumn

\counterwithin{figure}{section}
\counterwithin{table}{section}
\counterwithin{equation}{section}

\section{Reference Implementations}\label{sec:app:losses}

\begin{listing}[!ht]
\usemintedstyle{vs}
\begin{minted}{python}
import jax
import jax.scipy.special
import jax.numpy as jnp


def hl_gauss_transform(
    min_value: float,
    max_value: float,
    num_bins: int,
    sigma: float,
):
    """Histogram loss transform for a normal distribution."""
    support = jnp.linspace(min_value, max_value, num_bins + 1, dtype=jnp.float32)

    def transform_to_probs(target: jax.Array) -> jax.Array:
        cdf_evals = jax.scipy.special.erf((support - target) / (jnp.sqrt(2) * sigma))
        z = cdf_evals[-1] - cdf_evals[0]
        bin_probs = cdf_evals[1:] - cdf_evals[:-1]
        return bin_probs / z

    def transform_from_probs(probs: jax.Array) -> jax.Array:
        centers = (support[:-1] + support[1:]) / 2
        return jnp.sum(probs * centers)

    return transform_to_probs, transform_from_probs
\end{minted}
\vspace{-2.5mm}
\caption{An implementation of HL-Gauss \citep{imani2018improving} in Jax \citep{jax2018github}.}
\label{listing:losses:jax}
\end{listing}

\begin{listing}[!ht]
\usemintedstyle{vs}
\begin{minted}{python}
import torch
import torch.special
import torch.nn as nn
import torch.nn.functional as F


class HLGaussLoss(nn.Module):
    def __init__(self, min_value: float, max_value: float, num_bins: int, sigma: float):
        super().__init__()
        self.min_value = min_value
        self.max_value = max_value
        self.num_bins = num_bins
        self.sigma = sigma
        self.support = torch.linspace(
            min_value, max_value, num_bins + 1, dtype=torch.float32
        )

    def forward(self, logits: torch.Tensor, target: torch.Tensor) -> torch.Tensor:
        return F.cross_entropy(logits, self.transform_to_probs(target))

    def transform_to_probs(self, target: torch.Tensor) -> torch.Tensor:
        cdf_evals = torch.special.erf(
            (self.support - target.unsqueeze(-1))
            / (torch.sqrt(torch.tensor(2.0)) * self.sigma)
        )
        z = cdf_evals[..., -1] - cdf_evals[..., 0]
        bin_probs = cdf_evals[..., 1:] - cdf_evals[..., :-1]
        return bin_probs / z.unsqueeze(-1)

    def transform_from_probs(self, probs: torch.Tensor) -> torch.Tensor:
        centers = (self.support[:-1] + self.support[1:]) / 2
        return torch.sum(probs * centers, dim=-1)
\end{minted}
\vspace{-2.5mm}
\caption{An implementation of HL-Gauss \citep{imani2018improving} in PyTorch \citep{pytorch}.}
\label{listing:losses:torch}
\end{listing}

\clearpage

\section{Experimental Methodology}\label{app:exp}

In the subsequent sections we outline the experimental methodology for each domain herein.

\subsection{Atari}\label{app:exp:atari}

Both our online and offline RL regression baselines are built upon the Jax \citep{jax2018github} implementation of DQN+Adam in Dopamine \citep{castro18dopamine}.
Similarly, each of the classification methods (i.e., \hlgauss{} and \twohot{}) were built upon the Jax \citep{jax2018github} implementation of C51 in Dopamine \citep{castro18dopamine}.
Hyperparameters for DQN+Adam are provided in \autoref{table:atari:dqn} along with any hyperparameter differences for C51 (\autoref{table:atari:c51}), \twohot{} (\autoref{table:atari:c51}), and \hlgauss{} (\autoref{table:atari:hlgauss}).
Unless otherwise stated the online RL results in the paper were ran for 200M frames on 60 Atari games with five seeds per game. The offline RL results were ran on the 17 games in \citet{kumar2021implicit} with three seeds per game. The network architecture for both the online and offline results is the standard DQN Nature architecture that employs three convolutional layers followed by a single non-linear fully-connected layer before outputting the action-values.

\begin{table}[H]
    \begin{minipage}[t]{0.6\textwidth}
    \begin{threeparttable}
    \caption{\textbf{DQN+Adam Hyperparameters.}}
    \label{table:atari:dqn}
    \begin{tabularx}{0.975\textwidth}{ll}
    \toprule
Discount Factor $\gamma$ & 0.99 \\
$n$-step & 1 \\
Minimum Replay History & $20,000$ agent steps   \\
Agent Update Frequency & $4$ environment steps  \\
Target Network Update Frequency & $8,000$ agent steps \\
Exploration $\epsilon$ & $0.01$ \\
Exploration $\epsilon$ decay & $250,000$ agent steps \\
Optimizer & Adam \\
Learning Rate & $6.25 \times 10^{-5}$ \\
Adam $\epsilon$ & $1.5 \times 10^{-4}$ \\
Sticky Action Probability & $0.25$ \\
Maximum Steps per Episode & $27,000$ agent steps \\
Replay Buffer Size & $1,000,000$ \\
Batch Size & $32$ \\
    \bottomrule
    \end{tabularx}
    \end{threeparttable}
    \end{minipage}
    \hfill
    \begin{minipage}[t]{0.375\textwidth}
    \begin{threeparttable}
    \caption{\textbf{C51 \& \twohot{} Hyperparameters.} Difference in hyperparameters from DQN+Adam \autoref{table:atari:dqn}.}
    \label{table:atari:c51}
    \begin{tabularx}{\textwidth}{ll}
    \toprule
    Number of Locations & $51$ \\
    $[v_{\text{min}}, v_{\text{max}}]$ &  $[-10, 10]$ \\
    \midrule
    Learning Rate & $0.00025$ \\
    Adam $\epsilon$ & $0.0003125$ \\
    \bottomrule
    \end{tabularx}
    \end{threeparttable}
    \begin{threeparttable}
    \vspace{0.625cm}
    \caption{\textbf{HL-Gauss Hyperparameters.} Difference in hyperparameters from C51 \autoref{table:atari:c51}.}
    \label{table:atari:hlgauss}
    \begin{tabularx}{\textwidth}{ll}
    \toprule
    Smoothing Ratio $\sigma / \varsigma$ & 0.75 \\
    \bottomrule
    \end{tabularx}
    \end{threeparttable}
    \end{minipage}
\end{table}

\subsubsection{Mixtures of Experts}\label{app:exp:atari:moe}

All experiments ran with SoftMoE reused the experimental methodology of \citet{obando2024mixtures}. Specifically, we replace the penultimate layer of the DQN+Adam in Dopamine \citep{castro18dopamine} with a SoftMoE \citep{puigcerver2023sparse} module. The MoE results were ran with the Impala ResNet architecture \citep{espeholt2018impala}. We reuse the same set of $20$ games from \citet{obando2024mixtures} and run each configuration for five seeds per game.
All classification methods reused the parameters from \autoref{table:atari:c51} for C51 and \twohot{} or \autoref{table:atari:hlgauss} for \hlgauss{}.

\subsubsection{Multi-Task \& Multi-Game}

The multi-task and multi-game results follow exactly the methodology outlined in \citet{taiga2023multitask} and \citet{kumar2022offline} respectively.
We reuse the hyperparameters for HL-Gauss outlined in \autoref{table:atari:hlgauss}. For multi-task results each agent is run for five seeds per game. Due to the prohibitive compute of the multi-game setup we run each configuration for a single seed.

\subsection{Chess}\label{app:sec:chess}

We follow exactly the setup in \citet{ruoss2024grandmaster} with the only difference being the use of \hlgauss{} with a smoothing ratio $\sigma / \varsigma = 0.75$.
Specifically, we take the action-values produced by Stockfish and project them a categorical distribution using \hlgauss{}.
As \citet{ruoss2024grandmaster} was already performing classification we reuse the parameters of their categorical distribution, those being, $m = 128$ bins evenly divided between the range $[0, 1]$.
For each parameter configuration we train a single agent and report the evaluation puzzle accuracy.
Puzzle accuracy numbers for one-hot and AlphaZero w/ MCTS were taken directly from \citet[][Table 6]{ruoss2024grandmaster}.

\subsection{Robotic manipulation experiments.}
We study a large-scale vision-based robotic manipulation setting on a mobile manipulator robot with 7 degrees of freedom, which is visualized in Figure~\ref{fig:q_transformer} (left).
The tabletop robot manipulation domain consists of a tabletop with various randomized objects spawned on top of the countertop.
A RetinaGAN is applied to transform the simulation images closer to real-world image distributions, following the method in~\citep{ho2020retinagan}.
We implement a Q-Transformer policy following the procedures in~\citep{chebotar2023q}.
Specifically, we incorporate autoregressive $Q$-learning by learning $Q$-values per action dimension, incorporate conservative regularization to effectively learn from suboptimal data, and utilize Monte-Carlo returns.

\begin{figure}[H]
\centering
\includegraphics[width=0.925\linewidth]{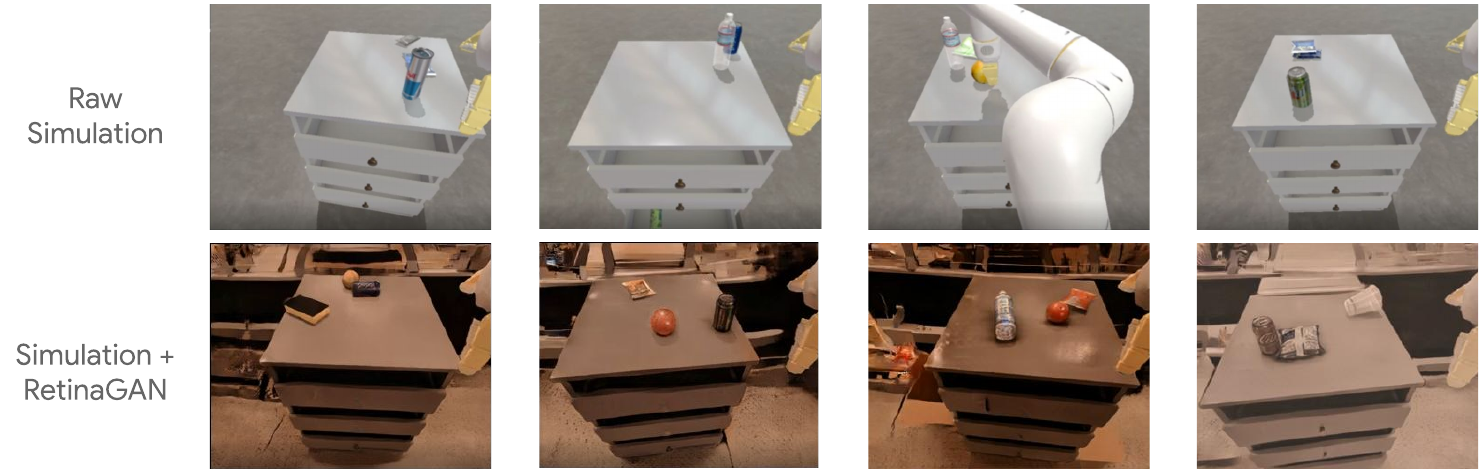}
\caption{\footnotesize{\textbf{Robot manipulation domain.} The simulated robot manipulation~(\sref{subsubsec:robot}) consists of a tabletop with randomized objects. A learned RetinaGAN transformation is applied to make the visual observation inputs more realistic.}}
\end{figure}

\subsection{Regression Target Magnitude \& Loss of Plasticity}
\label{sec:app}

To assess whether classification losses are more robust when learning non-stationary targets of increasing magnitude we leverage the synthetic setup from \citet{lyle2024disentangling}. Specifically, we train a convolutional neural network that takes CIFAR 10 images $x_i$ as input and outputs a scalar prediction: $f_\theta : \mathbb{R}^{32 \times 32 \times 3} \to \mathbb{R}$. The goal is to fit the regression target,
$$
y_i = \sin(m f_{\theta^{-}}(x_i)) + b \,
$$
where $m = 10^5$, $\theta^{-}$ are a set of randomly sampled target parameters for the same convolutional architecture, and $b$ is a bias that changes the magnitude of the prediction targets.
It is clear that increasing $b$ shouldn't result in a more challenging regression task.

When learning a value function with TD methods the regression targets are non-stationary and hopefully increasing in magnitude (corresponding to an improving policy). To simulate this setting we consider fitting the network $f_{\theta}$ on the increasing sequence $b \in \{0, 8, 16, 24, 32 \}$. For each value $b$ we sample a new set of target parameters $\theta^{-}$ and regress towards $y_i$ for $5,000$ gradient steps with a batch size of $512$ with the Adam optimizer using a learning rate of $10^{-3}$.

We evaluate the Mean-Squared Error (MSE) throughout training for three methods: \twohot{}, \hlgauss{}, and L2 regression. For both \twohot{} and \hlgauss{} we use a support of $[-40, 40]$ with $101$ bins. Figure~\ref{fig:synthetic_magnitude_main} depicts the MSE throughout training averaged over $30$ seeds for each method. One can see that the network trained with L2 regression does indeed loose its ability to rapidly fit targets of increasing magnitude, consistent with \citet{lyle2024disentangling}.
On the other hand, the classification methods are more robust and tend to converge to the same MSE irrespective of the target magnitude $b$. Furthermore, we can see that \hlgauss{} outperforms \twohot{}, consistent with our previous findings. These results help provide some evidence that perhaps one of the reasons classification outperforms regression is due to the network being more ``plastic'' under non-stationary targets.

\section{Per-Game Atari Results}

\begin{figure}[H]
\centering
\includegraphics[width=0.965\linewidth]{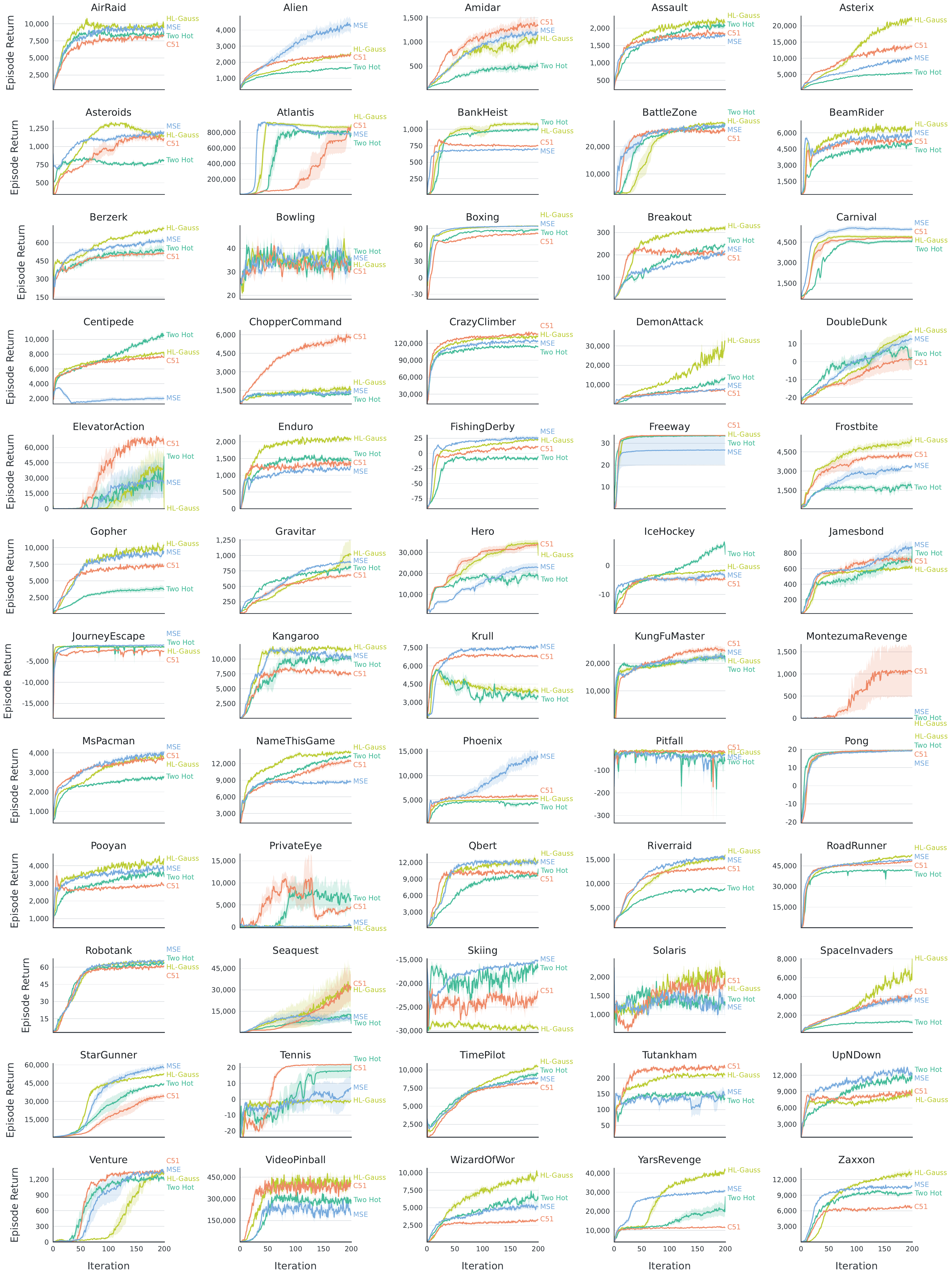}
\caption{\footnotesize{\textbf{Training curves on single-task online RL ~(\sref{subsec:single_task_atari}) for all 60 Atari games.} All games ran for 200M frames and ran for: DQN(Adam), C51, Two-Hot, and \hlgauss{}.}}
\label{fig:single-task-online-rl-training-curves}
\end{figure}

\begin{figure*}[!ht]
\centering
\begin{minipage}[c]{0.49\textwidth}
\centering
\includegraphics[width=0.98\linewidth]{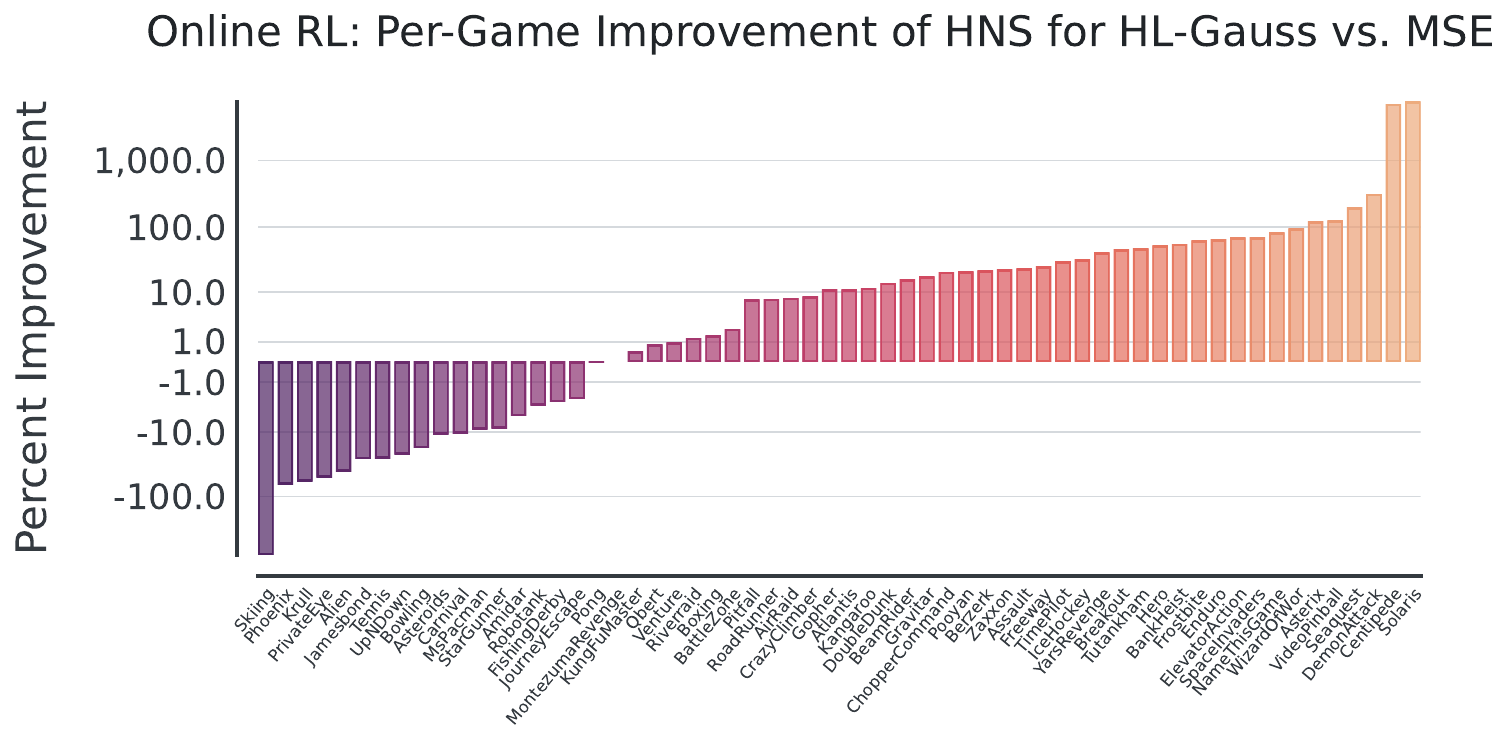}
\end{minipage}
\hfill
\begin{minipage}[c]{0.49\textwidth}
\centering
\includegraphics[width=0.98\linewidth]{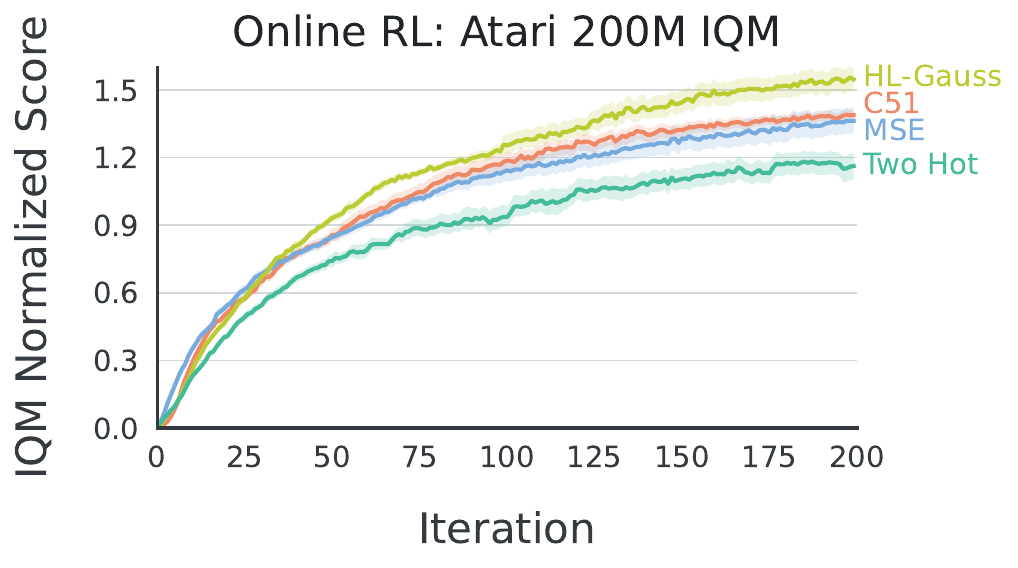}
\end{minipage}
\caption{\footnotesize{\textbf{\hlgauss{} vs MSE per game in single-task online RL  ~(\sref{subsubsec:multi-task})}. \textbf{(Left)} 
Each column displays the relative final performance of \hlgauss{} with respect to MSE in the single-task online RL training curves. This is a summary of the curves displayed in Figure~\ref{fig:single-task-online-rl-training-curves}. Note that \hlgauss{} outperforms MSE in $\approx3/4$ of all games, and that \hlgauss{} scores at least 10\% higher on 1/2 of all games. \textbf{(Right)} IQM normalized training curves throughout training.}} 
\label{fig:online-per-game-improvements}
\end{figure*}

\vspace{-5mm}
\section{Additional Results}

\begin{figure}[!ht]
\begin{minipage}[t]{0.49\textwidth}
\centering
\includegraphics[width=0.9\linewidth]{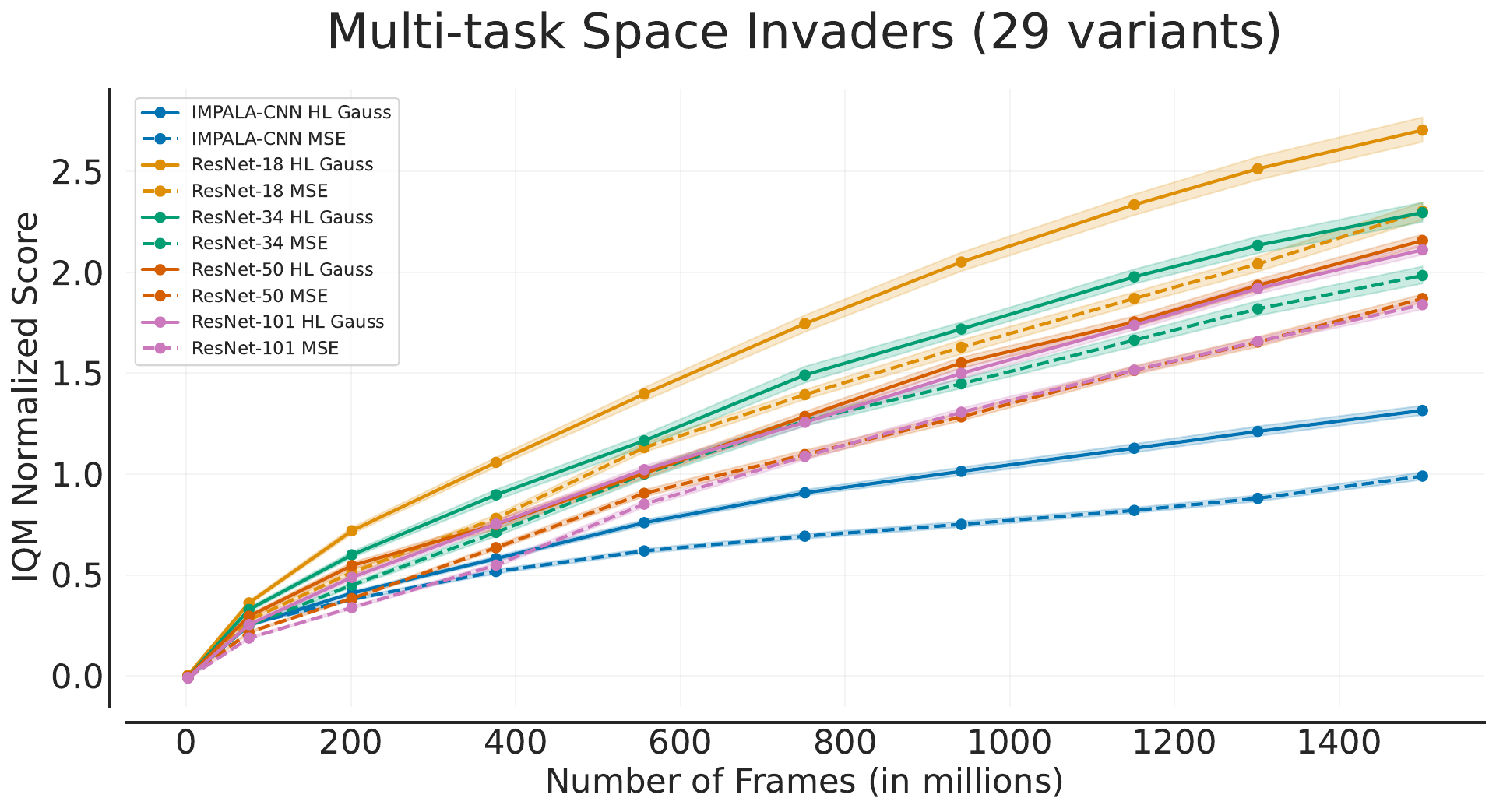}
\caption{\footnotesize{\textbf{Multi-task online RL ~(\sref{subsubsec:multi-task}) training curves for \textsc{Space Invaders} trained concurrently on 29 game variants.} Note that for every architecture, the \hlgauss{} variant scales better than its respective MSE variant.}}
\end{minipage}
\hfill
\begin{minipage}[t]{0.49\textwidth}
\centering
\includegraphics[width=0.9\linewidth]{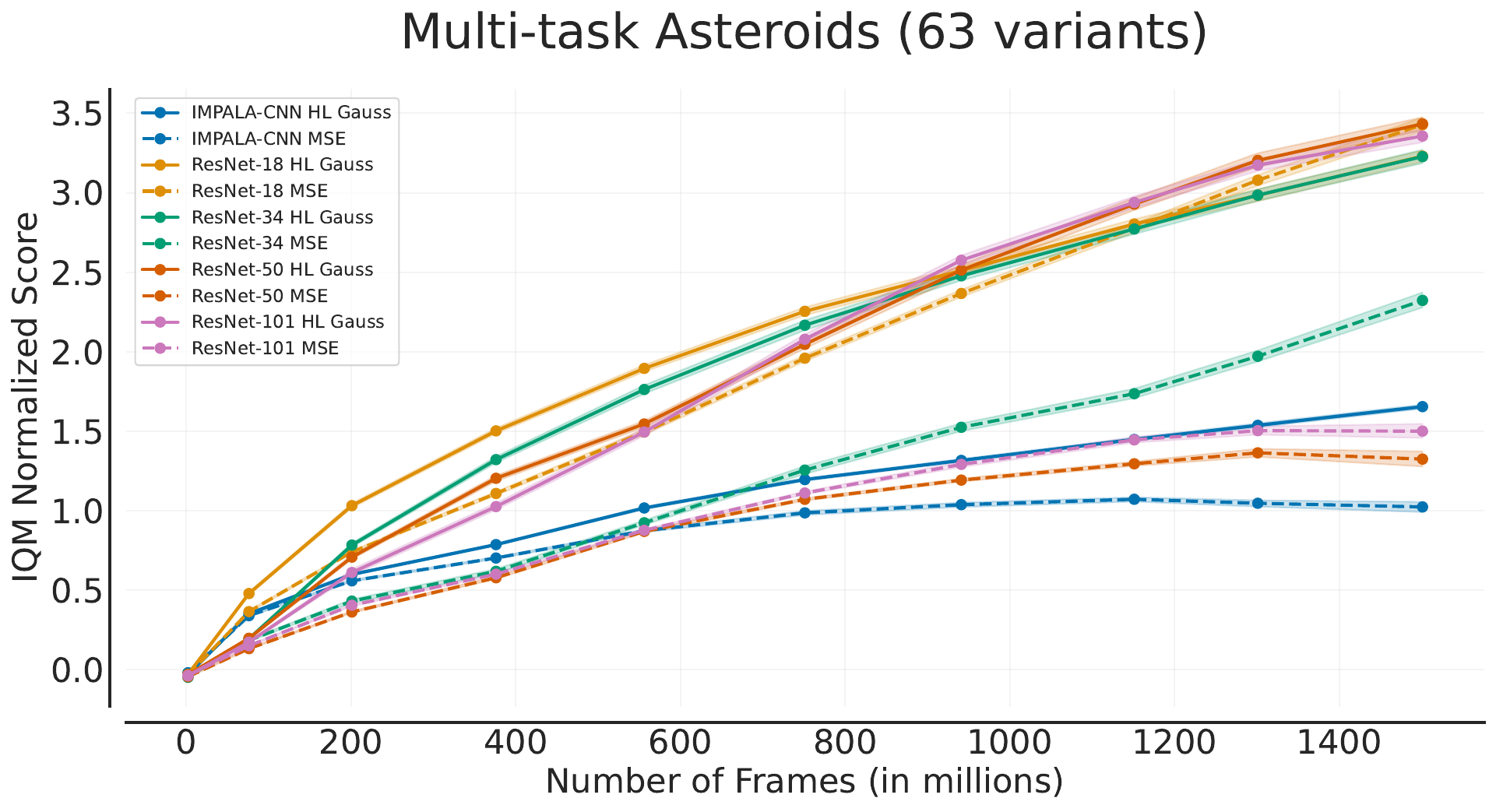}
\caption{\footnotesize{\textbf{Multi-task online RL ~(\sref{subsubsec:multi-task}) training curves for \textsc{Asteroids} trained concurrently on 63 game variants}. These results investigate the scaling properties per architecture of MSE critic loss and cross-entropy \hlgauss{} loss. Note that with the architectures larger than Resnet 18, \hlgauss{} keeps improving while MSE performance drops after 1300M frames. These larger architectures also all reach higher peak IQM scores with \hlgauss{}.}}
\end{minipage}
\end{figure}

\begin{figure*}[!ht]
\centering
\begin{minipage}[t]{0.49\textwidth}
\includegraphics[width=\linewidth]{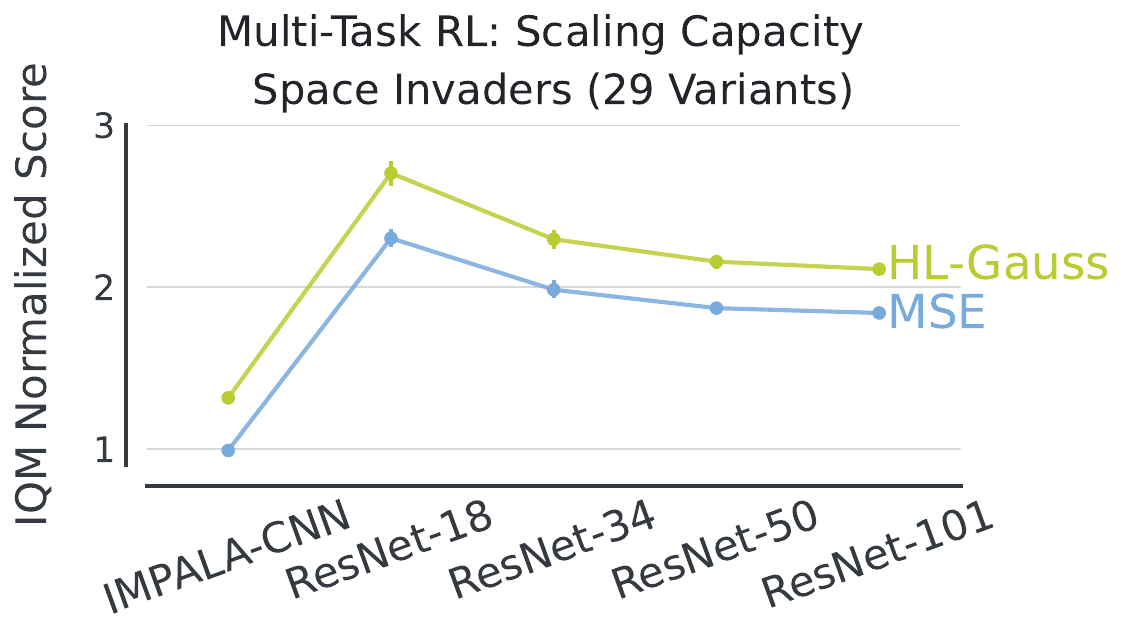}
\caption{\footnotesize{\textbf{Scaling curves on Multi-task Online RL}. %
Online RL scaling results with actor-critic IMPALA with ResNets on \textsc{Space Invaders}. HL-Gauss outperforms MSE for all models. Since human scores are not available for  variants, we report normalized scores using a baseline IMPALA agent with MSE loss. See \sref{subsubsec:multi-task} for more details.}}
\label{fig:online_scaling_with_spaceinvaders}
\end{minipage}
\hfill
\begin{minipage}[t]{0.49\textwidth}
\centering
\includegraphics[width=\linewidth]{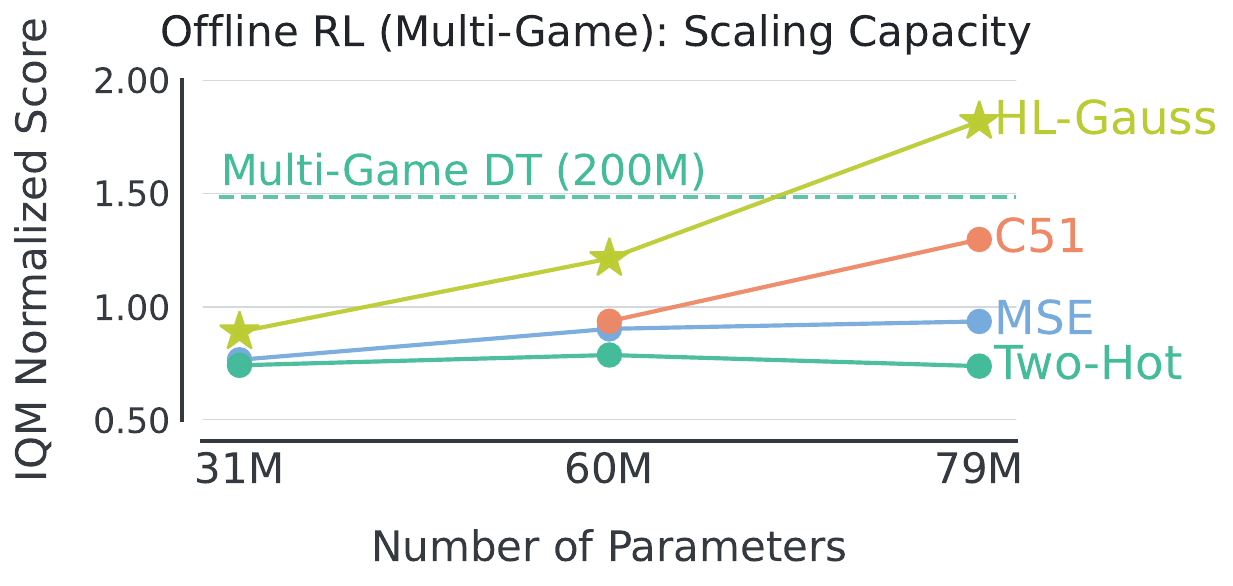}
\caption{\footnotesize{\textbf{Multi-task Offline RL results presented in terms of DQN normalized scores}. Note that when aggregate results are computed with DQN normalization, \hlgauss{} exhibits a faster rate of improvement than C51 as the number of parameters scales up.}}
\label{fig:offline_scaling_dqn_norm}
\end{minipage}
\end{figure*}

\end{document}